\documentclass[10pt,twocolumn,letterpaper]{article}

\usepackage[pagenumbers]{cvpr} 

\usepackage{appendix}
\usepackage{multicol}
\usepackage{multirow}
\usepackage{xcolor}
\usepackage{graphicx} 
\usepackage{caption}
\usepackage{subcaption} 
\usepackage{float} 

\definecolor{cvprblue}{rgb}{0.21,0.49,0.74}
\usepackage[pagebackref,breaklinks,colorlinks,allcolors=cvprblue]{hyperref}

\def\paperID{21266} 
\def\confName{CVPR}
\def\confYear{2026}

\title{Age-Inclusive 3D Human Mesh Recovery for Action-Preserving \\ Data Anonymization}

\author{
Georgios Chatzichristodoulou$^{1,2}$ \quad
Niki Efthymiou$^{2,3}$ \quad
Panagiotis Filntisis$^{2,3}$ \\ \quad
Georgios Pavlakos$^4$ \quad
Petros Maragos$^{1,2,3}$
\vspace{0.2cm}\\
$^{1}$School of ECE, National Technical University of Athens \quad
$^{2}$Robotics Institute, Athena RC\\
$^{3}$HERON - Hellenic Robotics Center of Excellence \quad
$^{4}$The University of Texas at Austin\\
}

\begin{document}

\twocolumn[{%
\renewcommand\twocolumn[1][]{#1}%
\maketitle
\begin{center}
    \vspace{-0.26in}
    \centerline{
   \includegraphics[width=0.85\textwidth,clip]{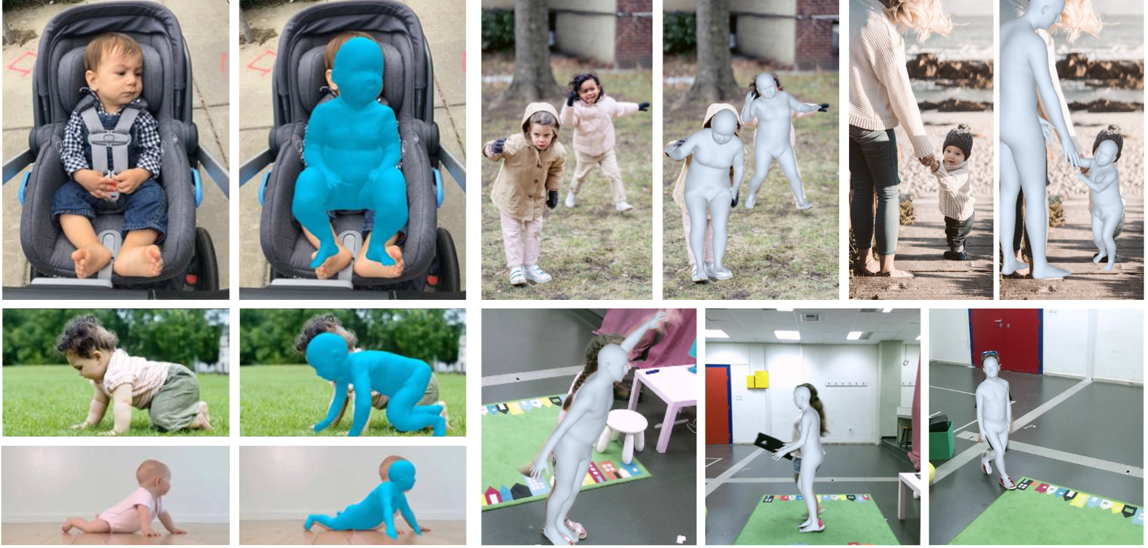}
     }
    \vspace{-0.05in}
   \captionof{figure}{\textbf{Age-Inclusive Human Mesh Recovery with AionHMR}. We present \textbf{AionHMR}, a novel framework for age-inclusive human mesh recovery from single RGB images. The framework consists of two stages: \textbf{AionHMR-a}, an optimization-based method that estimates SMPL-A shape and pose parameters along with camera parameters; and \textbf{AionHMR-b}, a transformer-based, real-time HMR network trained using pseudo-ground-truth annotations generated by AionHMR-a.  We display AionHMR-a fittings (left, blue) and AionHMR-b results (right, gray). The bottom right shows 3D-BabyRobot database samples of child-robot interaction reconstructions.
   }
   \vspace{-0.02in}
\label{fig:teaser}
\end{center}%
}]

\begin{abstract}
While three-dimensional (3D) shape and pose estimation is a highly researched area that has yielded significant advances, the resulting methods, despite performing well for the adult population, generally fail to generalize effectively to children and infants. This paper addresses this challenge by introducing AionHMR, a comprehensive framework designed to bridge this domain gap. We propose an optimization-based method that extends a top-performing model by incorporating the SMPL-A body model, enabling the concurrent and accurate modeling of adults, children, and infants. Leveraging this approach, we generated pseudo-ground-truth annotations for publicly available child and infant image databases. Using these new training data, we then developed and trained a specialized transformer-based deep learning model capable of real-time 3D age-inclusive human reconstruction. Extensive experiments demonstrate that our methods significantly improve shape and pose estimation for children and infants without compromising accuracy on adults. Importantly, our reconstructed meshes serve as privacy-preserving substitutes for raw images, retaining essential action, pose, and geometry information while enabling anonymized datasets release. As a demonstration, we introduce the 3D-BabyRobot dataset, a collection of action-preserving 3D reconstructions of children interacting with robots. This work bridges a crucial domain gap and establishes a foundation for inclusive, privacy-aware, and age-diverse 3D human modeling.
\end{abstract}

\vspace{-0.5cm}
\section{Introduction}
\label{sec:intro}

Estimating 3D human shape and pose from a single image is a core Computer Vision task with applications in health~\cite{kojovic2021using, ganai2025early}, sports~\cite{baumgartner2023monocular}, VR~\cite{belghit2018vision}, and animation~\cite{Roussos2025three}. Recent advances in statistical body models and deep learning have led to strong performance for adults, supporting progress in action recognition~\cite{rajasegaran2023benefits}, gait analysis~\cite{9881695}, and HRI planning~\cite{10.1007/978-981-99-6498-7_16}.

Although most modern 3D shape and pose estimation methods have been highly successful for adults ~\cite{goel2023humans,kolotouros2021probabilistic}, extending them to younger populations, such as children and babies, remains largely unresolved. The core obstacles are both ethical and technical, creating a unique research gap centered on data scarcity: existing human body models like SMPL~\cite{SMPL:2015} are built from thousands of adult 3D scans, while strict legislation and ethical complexities make acquiring comparable child data exceptionally difficult. This creates a considerable domain gap, as adult-based models fail to capture the dramatic anthropometric changes from infancy through childhood to adulthood-variations crucial for accurate modeling. This lack of public data is further exacerbated by inaccessible sensitive datasets (\eg, real child or patient data), preventing the development of more accurate pediatric models. While models like SMIL~\cite{10.1007/978-3-030-00928-1_89} and SMPL-Agora (SMPL-A)~\cite{Patel:CVPR:2021} better represent infant and child bodies, fitting them to images remains challenging.

Motivated by the above, we propose \textbf{AionHMR (Age Inclusive Optimization and Network for Human Mesh Recovery)}, a unified framework for human mesh recovery from single RGB images across all age groups. It operates by generating pseudo-ground-truth annotations from these images, which are then used to create datasets to train a model that regresses the shape and pose parameters of a human body model. Our approach comprises two complementary methods both supporting single and multi-person scenarios and specifically designed to better model children and babies, while they can be universally applied to humans of all ages.

The first method, AionHMR-a, is an optimization-based approach that fits the SMPL-A human body model by minimizing an objective function combining 3D reprojected keypoint alignment with 2D detections, regularization terms, and improved detection and tracking strategies. While effective for generating high-quality 3D meshes, its iterative nature makes it slow and susceptible to local minima; hence, its primary use is creating pseudo-ground-truth annotations to train a faster learning-based model. The second method, AionHMR-b, is a learning-based approach that predicts 3D shape and pose from a single image with much faster inference but requires large-scale annotated datasets. To address the lack of child-specific annotations, we use AionHMR-a to generate high-quality pseudo-ground-truth labels, effectively augmenting existing datasets.

To validate our approach and demonstrate its practical relevance, we introduce 3D-BabyRobot, a novel dataset featuring 3D reconstructions of children interacting with robots. The dataset establishes a framework for the ethical collection and anonymization of sensitive visual data, where 3D body representations replace identifiable imagery while preserving essential information about human pose, geometry, and behavior.

Our contributions address the critical data and methodological limitations in 3D shape and pose estimation for non-adult subjects:

\begin{itemize}
 \item We adapt an optimization-based technique for 3D shape and pose estimation, specifically to enable robust and accurate reconstruction for infants and young children. This allows us to generate training data by curating image datasets with 3D pseudo-ground-truth annotations. We release these annotations as a valuable resource to the community.
 \item  We introduce a specialized, highly accurate HMR-like deep learning model for 3D shape and pose estimation. Built by integrating a pre-trained backbone with a customized SMPL-A head and trained on 3D pseudo-ground-truth generated by our optimization-based method, it achieves markedly better performance on child and infant imagery than standard adult-trained or age-inclusive HMR models.
\item We publicly release the 3D-BabyRobot dataset to enable future research in child-robot interaction analysis while maintaining ethical data sharing standards.\footnote{AionHMR's code and 3D BabyRobot Dataset are available at: \texttt{https://github.com/gioxatz/aionhmr}} 

\end{itemize}

\vspace{-0.3cm}
\section{Related Work}
\begin{figure*}[t]
    \centering
    \includegraphics[width=1\linewidth]{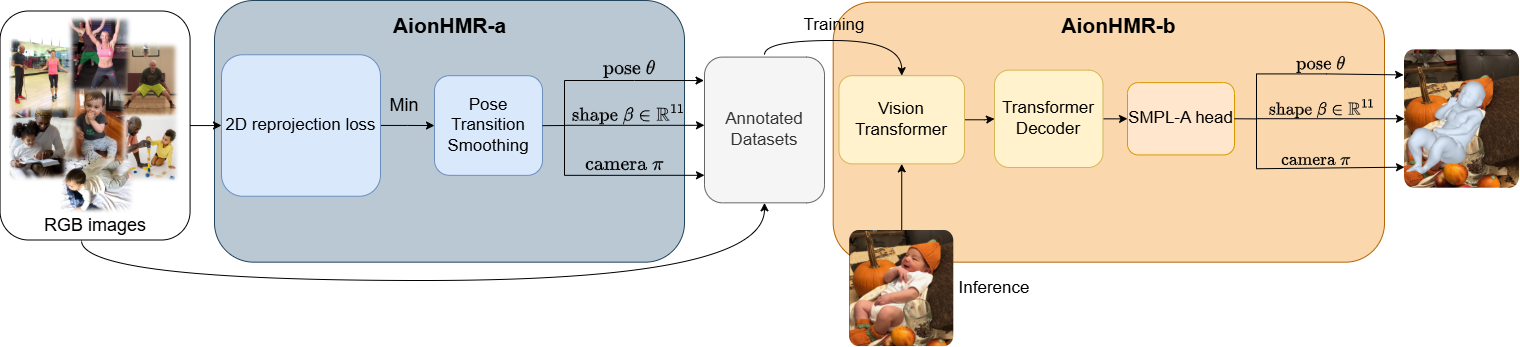}
    \caption{\textbf{AionHMR pipeline}. AionHMR-a is an optimization-based technique for the estimation of SMPL-A shape and pose parameters, and camera parameters based on a 2D reprojection loss of the keypoints. We use AionHMR-a to annotate images and create datasets. Using these datasets, we train a Transformer-based model, AionHMR-b, that regresses in real-time the same SMPL-A and camera parameters.}
    \label{fig:pipeline}
\end{figure*}

The estimation of 3D human shape and pose from a single image or video is a central research problem. As detailed shape reconstruction inherently requires pose information, most contemporary methods jointly estimate both.

\vspace{-0.2cm}
\subsection{3D Shape and Pose Estimation} 

Early approaches to 3D pose estimation often lifted 2D joint estimates (frequently from CNNs) to 3D coordinates \cite{chen20173dhumanposeestimation, martinez2017simple, moreno20173d, park20163dhumanposeestimation}. Subsequent end-to-end methods were developed to bypass this intermediate 2D step \cite{motionagformer2024, liao2023multipleviewgeometrytransformers}, with some exploring novel inputs like WiFi signals \cite{person3dyan}.

To address inherent depth and shape ambiguities, many methods leverage richer data sources. These include multi-view cameras \cite{Hofmann2012, dong2021shapeawaremultipersonposeestimation, pavlakos2017harvestingmultipleviewsmarkerless}, video sequences \cite{ye2023slahmr, rajasegaran2022tracking}, multi-shot still images \cite{pavlakos2022multishot}, or RGB-D sensors \cite{bashirov2021realtimergbdbasedextendedbody}. Analyzing subjects from multiple viewpoints helps resolve ambiguities and improve accuracy. More recently, models have integrated biomechanical constraints to generate more plausible motion and reject physically impossible poses \cite{xia2025reconstructinghumansbiomechanicallyaccurate, koleini2025bioposebiomechanicallyaccurate3dpose}.

Reconstruction from a single RGB image remains the most challenging scenario due to obscured depth. Methods addressing this problem can be broadly categorized as optimization-based \cite{Bogo:ECCV:2016, SMPL-X:2019, fan2021revitalizingoptimization3dhuman}, regression-based using deep learning \cite{dwivedi_cvpr2024_tokenhmr, hmrKanazawa17, goel2023humans, Zheng2019DeepHuman, BEV}, or hybrid approaches that combine both \cite{kolotouros2019spin}.

\vspace{-0.2cm}

\subsection{3D Child Shape and Pose Estimation}
\vspace{-0.15cm}
Most methods that estimate 3D human shape and pose are typically trained on adult datasets and thus fail to generalize effectively to children and infants. In order to model children and infants two additional models have been proposed, SMIL~\cite{10.1007/978-3-030-00928-1_89} and  SMPL-A~\cite{Patel:CVPR:2021}. SMIL is a specialized, learned model for infant human body reconstruction.  By focusing on infants, SMIL overcomes the inherent biases of models trained on adult data. SMPL-A creates a mesh template by interpolating a SMPL and a SMIL mesh template, enabling it to model infants, children and adults simultaneously with one model.  
\vspace{-0.6cm}
\paragraph{SMIL} SMIL is trained on a unique RGB-D dataset of freely moving infants, addressing the real-world challenge of collecting data from uncooperative subjects. Despite the low-quality inputs, the model effectively learns robust representations of dynamic, unconstrained infant motions. Built on the SMPL framework, SMIL employs a template mesh driven by shape and pose parameters, specifically optimized to capture infant anatomy, including distinctive bone structure and fat distribution.
\vspace{-0.5cm}

\paragraph{Methods} 

Methods for child and infant mesh recovery remain scarce in the literature. BEV~\cite{BEV} is one method that addresses this challenge in multi-person scenarios by employing a unified 3D body model capable of handling large age variations. To jointly model adults and children, BEV adopts a modified SMPL-A model that integrates SMPL and SMIL. When the interpolation parameter $\alpha > t_\alpha$, the SMIL model is used to represent infants better, as their body shape diverges from typical child and adult shapes. Otherwise, for $\alpha \leq t_\alpha$, the standard SMPL model is applied with $\alpha$ controlling the shape interpolation between adult ($\alpha=0$) and child ($\alpha=1$) templates. 

\vspace{-0.25cm}

\section{Preliminaries}
\vspace{-0.2cm}

\paragraph{SMPL} SMPL~\cite{SMPL:2015} is a parametric human body model. Starting from a template mesh, deformations controlled by shape parameters $\beta \in \mathbb{R}^{10}$ and pose parameters $\theta \in \mathbb{R}^{72}$ generate a mesh that fits the human body. It is widely used in 3D human modeling but was trained on adult body scans only and thus does not generalize well to children or infants.
\vspace{-0.9cm}
\paragraph{SMPL-A} To model adults, children, and infants concurrently, we use the SMPL-A model. The key difference from SMPL lies in the body template: SMPL-A interpolates between an adult template $T_A$ (from SMPL) and a child template $T_C$ (from SMIL) using a shape interpolation parameter $\alpha \in [0,1]$:   
$\alpha T_{C} + (1 - \alpha)T_{A}$.
Apart from the template, SMPL-A retains the adult shape space of SMPL, making the body template the main distinction. Generally, as well as in our paper, we set $\alpha$ as the $11^{th}$ shape $\beta$ parameter. 
\vspace{-0.5cm}
\paragraph{HMR} 
  Human Mesh Recovery methods train a predictor that, given an input image, regresses the shape and pose parameters of a parametric human body model to generate a 3D mesh that aligns with the depicted person. The predictor is usually implemented as a deep neural network and trained on large-scale datasets to capture diverse human shapes and poses, improving generalization to in-the-wild images.

\vspace{-0.3cm}

\section{Method}

We introduce \textbf{AionHMR}, a unified framework for age-inclusive human mesh recovery. As shown in Figure~\ref{fig:pipeline}, it comprises two complementary stages. AionHMR-a is an optimization-based procedure, adapted from SLAHMR~\cite{ye2023slahmr} parts and equipped with the SMPL-A body model, that delivers highly accurate reconstructions when latency is not critical. Its outputs serve as high-quality pseudo-ground-truth annotations used to train AionHMR-b, a data-driven regressor that enables fast inference while retaining accuracy across all age groups.

\subsection{AionHMR-a: Optimization-based method}
The first part of AionHMR is an optimization-based method that estimates a person's 3D shape and pose. AionHMR-a is based on SLAHMR but utilizes a subset of its original components in our optimization process. SLAHMR jointly optimizes the 3D shape and pose of the humans in a video, as well as the camera motion. The observed motion of a person in camera coordinates depends on both the human's movement and the camera's movement relative to the world coordinates. Therefore, accurately modeling the camera motion is essential for the correct estimation of human motion. 

\vspace{-0.6cm}
\paragraph{Body Model} The first important change over SLAHMR involves the human body model: we switch from the SMPL+H model~\cite{MANO:SIGGRAPHASIA:2017} to the SMPL-A model. This modification is critical as it enables the estimation and modeling of all human ages.
Additionally, we enhance the pipeline for AionHMR-a to expand its input capability. While SLAHMR was restricted to video input, our system is modified to efficiently process both images and video. 

\noindent
\textbf{Optimization Stages}\hspace{2mm} The first step is to estimate each person's per-frame pose and compute their unique identity track associations over all frames using 4DHumans~\cite{goel2023humans} 3D tracking system.
 The optimization process consists of two phases:
The first one optimizes root translation and global orientation based on a 2D reprojection loss that aligns 3D projected joints with 2D detections from ViTPose~\cite{xu2022vitpose}.
The second phase optimization aims to smooth the transitions between poses across different frames. To this end, priors for joint smoothness, shape and pose plausibility are added. Here, the optimization is also performed on the shape and pose parameters, as well as the camera scale.

\noindent
\textbf{Initialization}\hspace{2mm} Since our primary focus is on children and babies, we initialized $\alpha$ at 1 (only child template) instead of 0 as originally, allowing the optimization to proceed from a child-centered starting point. This modification consistently improved the quality of the reconstruction compared to previous configurations, like grid search on the $\alpha$ value, which, despite the high-quality results, was time-ineffective.

\vspace{-0.25cm}
\subsection{AionHMR-b: Transformer-based HMR}

The high quality of the results of AionHMR-a in modeling children and babies, combined with the scarcity of child-specific annotated data available for training HMR-like models, encouraged us to create specialized datasets. Therefore, we extend an HMR-like model architecture and we train a model on both children and adult data.

\noindent
\textbf{Architecture}\hspace{2mm} More specifically, our base method is HMR2.0~\cite{goel2023humans}, a simple end-to-end transformer-based HMR approach. An image is passed through a ViT backbone~\cite{xu2022vitpose} that has been pre-trained on the task of 2D keypoint detection and produces the image tokens. These tokens are the input of a transformer decoder~\cite{NIPS2017_3f5ee243} with multi-head self-attention that feeds a SMPL-A head, which regresses the SMPL-A shape $\beta$ and pose $\theta$ parameters, as well as the camera parameters $\pi$. As we explained, $\beta\in\mathbb{R}^{11}$, where the $11^{th}$ value is the $\alpha$ interpolation weight.

\noindent
\textbf{Losses}\hspace{2mm}The losses used depend on the ground-truth or pseudo-ground-truth annotations contained in the datasets used to train the model.
Thus, an L2 norm is used for the SMPL-A parameters ($\mathcal{L}_{\texttt{smpl}})$. For keypoint supervision, an L1 norm is applied in two domains: directly on the predicted 3D keypoints ($\mathcal{L}_{\texttt{kp3D}}$) and on the corresponding 2D keypoints projected from 3D ($\mathcal{L}_{\texttt{kp2D}}$). Finally, to get plausible 3D poses, a discriminator $D_k$ is trained for each factor of the body model ($\mathcal{L}_{\texttt{adv}}$).

\noindent
\textbf{Training AionHMR-b}\hspace{2mm} Due to the unsatisfactory results obtained from more straightforward training and fine-tuning approaches, we adopted a combined training and hybrid model strategy. Initially, we trained a model from scratch, utilizing both the original HMR2.0 training datasets and our custom datasets. While this model learned to estimate the 3D shape successfully, we observed persistent issues with pose estimation. Testing the original HMR2.0 checkpoint on the same images revealed its superior pose accuracy. Consequently, we devised a hybrid model: we combined the SMPL-A head and the discriminator from our newly trained model with the ViT backbone derived from the original HMR2.0 checkpoint. Despite the immediate improvement in results, this new hybrid configuration requires further fine-tuning for a few epochs to ensure optimal adaptation and alignment between the distinct components. Following this fine-tuning, the model achieved the best results across all previous experiments, demonstrating high accuracy in both 3D shape and pose estimation. 
  \vspace{-0.2cm}

\section{Experiments}
\label{sec:exp}

\begin{table*}[h!t]
\centering
 \caption{\textbf{3D Shape Estimation - AHD and APHD metrics} (lower is better). Best results per metric and dataset are shown in \textbf{bold}. AionHMR clearly estimates the best 3D shape, while the other methods overestimate the height of the subjects, in order to fit the visual observations correctly.}
\label{tab:evaluation_results_single}
\setlength{\tabcolsep}{3.5pt} 
\renewcommand{\arraystretch}{1.1}
\resizebox{\textwidth}{!}{ 
\small
\scalebox{0.8}{
\begin{tabular}{l c c c c c c c c}
\toprule
\textbf{Dataset} & \multicolumn{2}{c}{\textbf{SyRIP}} & \multicolumn{2}{c}{\textbf{Relative Human}} & \multicolumn{2}{c}{\textbf{ChildPlay}} & \multicolumn{2}{c}{\textbf{BabyRobot}} \\
\cmidrule(lr){1-1} \cmidrule(lr){2-3} \cmidrule(lr){4-5} \cmidrule(lr){6-7} \cmidrule(lr){8-9}
\textbf{Method} & AHD $\downarrow$ (m) & APHD $\downarrow$ (\%) & AHD $\downarrow$ (m) & APHD $\downarrow$ (\%) & AHD $\downarrow$ (m) & APHD $\downarrow$ (\%) & AHD $\downarrow$ (m) & APHD $\downarrow$ (\%) \\
\midrule
ProHMR~\cite{kolotouros2021probabilistic} & -1.011 & -161.76 & -0.098 & -17.73 & -0.477 & -92.01 & -0.562 & -56.69 \\
HMR2.0~\cite{goel2023humans} & -0.980 & -157.62 & -0.075 & -16.18 & -0.468 & -90.80 & -0.534 & -54.27 \\
TokenHMR~\cite{dwivedi_cvpr2024_tokenhmr} & -0.976 & -156.34 & -0.091 & -17.21 & -0.441 & -88.70 & -0.562 & -56.74 \\
BEV~\cite{BEV} & -0.528 & -91.80 & \textbf{-0.009} & -12.50 & -0.067 & \textbf{-42.80} & -0.359 & -38.92 \\ \midrule
\textbf{AionHMR} & \textbf{0.022} & \textbf{-5.14} & 0.120 & \textbf{-2.21} & \textbf{-0.059} & -45.56 & \textbf{-0.283} & \textbf{-30.20} \\
\bottomrule
\end{tabular}
}
}
\vspace{-0.5cm}
\end{table*}

\subsection{Implementation Details}
For the code of the methods we use PyTorch~\cite{10.5555/3454287.3455008}.
For AionHMR-a, both stages are optimized with SLAHMR settings and the L-BFGS algorithm with a learning rate of 1.
For both the training and fine-tuning of the AionHMR-b model, we use the same configuration. The batch size is set to 16, we use AdamW optimizer~\cite{loshchilov2017decoupled} with a learning rate of $4\cdot10^{-5}$, $\beta_1 = 0.9, \beta_2=0.999$ and a weight decay of $10^{-4}$. The first training phase lasts for 2.5M steps, while the fine-tuning phase lasts 1.5M steps. For the different weights used during the training of AionHMR-b we set the values to $\mathcal{L}_\text{kp3D} = 0.05, \mathcal{L}_\text{kp2D} = 0.05, \mathcal{L}_\text{adv} = 0.0005$ and the terms within $\mathcal{L}_\text{smpl}$ weigh $0.0015$ and $0.001$ for the $\beta$ and $\theta$, respectively.

\subsection{Setup}

\paragraph{Training datasets}
For the training, we use the mixture of datasets used in the HMR2.0 model training, and, additionally, we use datasets that contain child and infant data. 
Specifically, we use the SyRIP~\cite{huang2021infant} (infants) and the Relative Human~\cite{BEV} (mixed ages) datasets with pseudo-ground truth SMPL parameters from AionHMR-a and 2D keypoints from their annotations, which were corrected (using ViTPose keypoints) when erroneous. Due to issues with extreme poses and unclear faces, we excluded low-quality samples. 
To ensure reliable facial information, crucial for distinguishing infants from adults, we limited training data to samples where ViTPose face keypoints confidence exceeded 0.7. Due to the data uniqueness and scarcity, we release these annotated samples.
During training, we prioritize these custom-annotated samples by assigning them a greater sampling probability, reflecting our main focus on child and baby data.

\vspace{-0.5cm}
\paragraph{Evaluation datasets}
Most evaluation datasets in this field either lack children and infant subjects or only test 3D and 2D poses. To evaluate the performance of different methods in 3D shape estimation, we generated pseudo-ground-truth annotations for images containing mainly infants and children using our optimization method. The images are from the SyRIP, Relative Human, Childplay~\cite{tafasca2023childplay} and BabyRobot~\cite{efthymiou2022childbot}, a multi-view RGB-D dataset featuring children interacting with robots. Of the four RGB-D views (front, sides, and top), we use images from the front and sides cameras for evaluation and the depth information of the top camera for the ground-truth height estimation. Finally, we evaluate AionHMR on COCO~\cite{lin2014microsoft} to test its efficacy on adults and a greater age range.

\vspace{-0.5cm}
\paragraph{Metrics}
For the evaluation, we use the \textbf{Mean Per Joint Position Error (MPJPE)} and the \textbf{Percentage of Correct Keypoints (PCK)} metrics for 3D pose and 2D pose, respectively.
Furthermore, we report the average mesh height generated by each method and introduce two quantitative metrics based on height: \textbf{Average Height Difference (AHD)} between predicted height and the pseudo-ground-truth calculated by AionHMR-a, and \textbf{Average Percentage Height Difference (APHD)}, calculated as a percentage of the ground truth. Let $N$ denote the total number of subjects, $H^*_i$ be the pseudo-ground-truth height (in centimeters), and $H_i$ be the predicted height for subject $i$. We calculate APHD as follows:
\vspace{-0.4cm}

\begin{equation}
\text{APHD} = \frac{100\%}{N}\cdot\sum_{i=1}^N\frac{H^*_i-H_i}{H^*_i}
\end{equation}

\vspace{-0.2cm}
Absolute values are omitted to reveal systemic bias, whether the model consistently over-predicts (positive) or under-predicts (negative) height.
\subsection{Results}
\begin{figure*}[!t] 
    \centering
  
    \begin{subfigure}[b]{0.3\textwidth} 
        \centering
        \includegraphics[width=\linewidth, trim=0mm 1mm 0mm 2mm, clip]{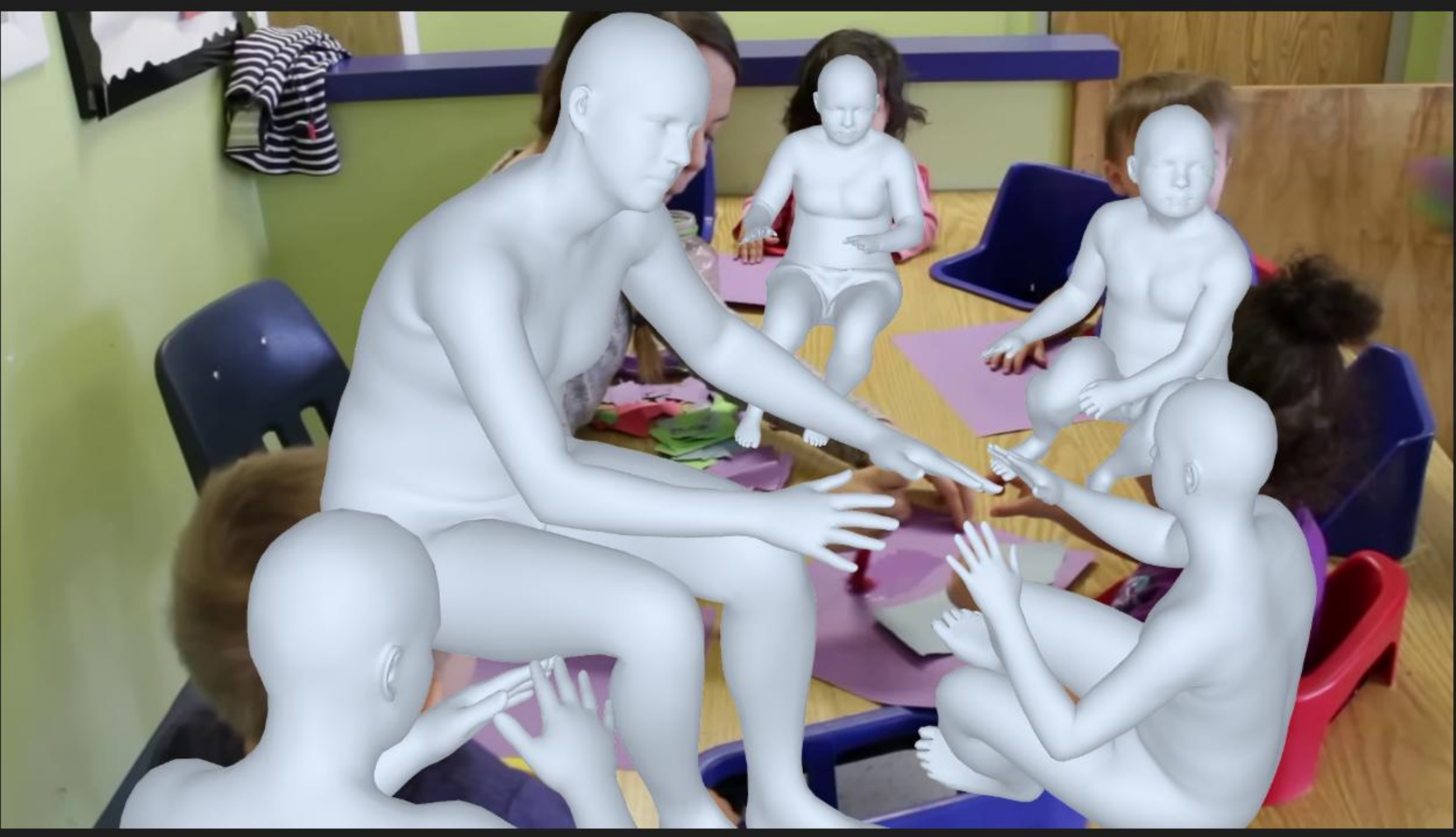}
        \caption{AionHMR}
        \label{fig:comparison:mine}
    \end{subfigure}
    \hfill 
    \begin{subfigure}[b]{0.3\textwidth}
        \centering
        \includegraphics[width=\linewidth]{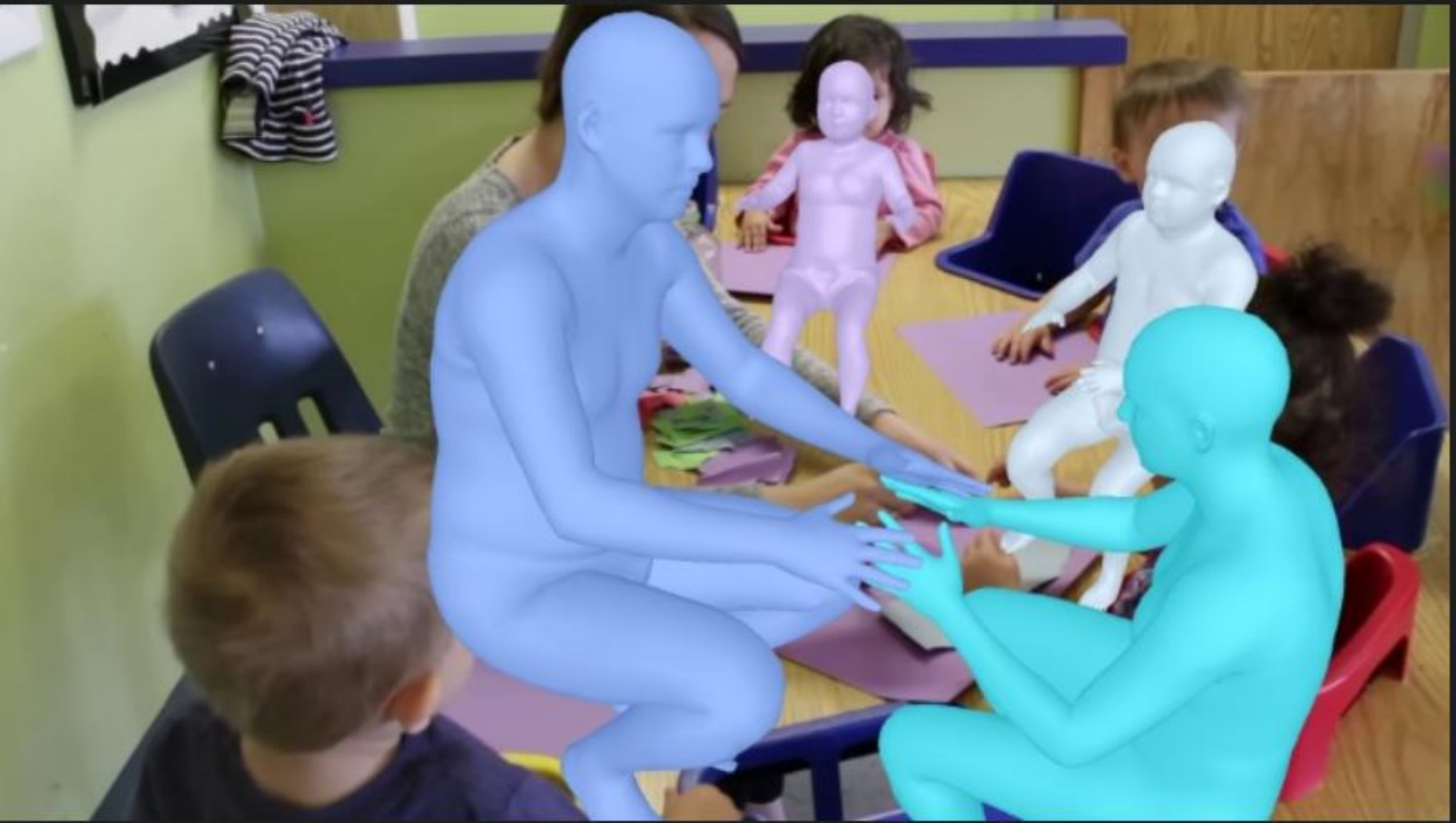}
        \caption{BEV}
        \label{fig:comparison:bev}
    \end{subfigure}
    \hfill
    \begin{subfigure}[b]{0.3\textwidth}
        \centering
        \includegraphics[width=\linewidth]{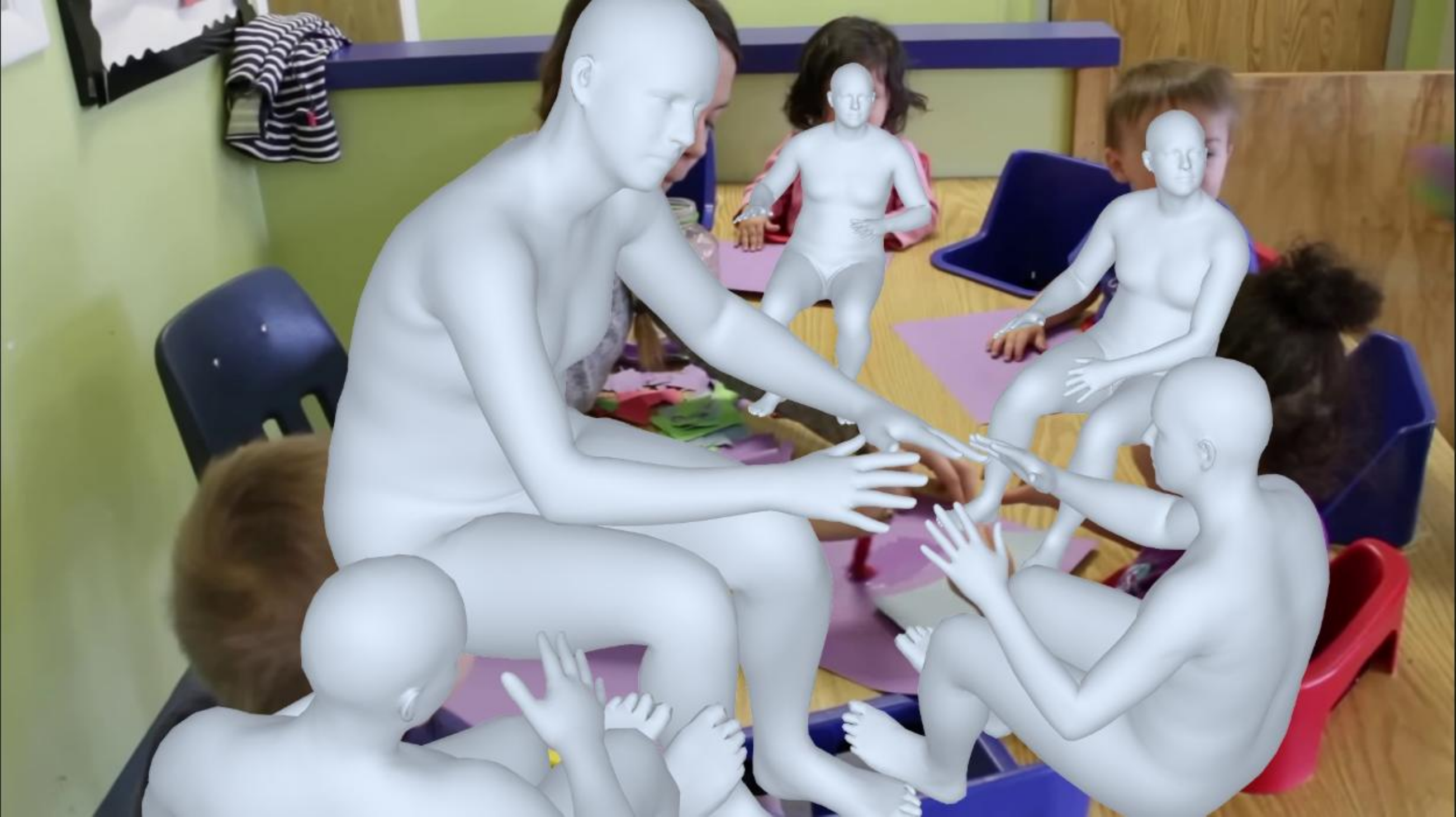}
        \caption{HMR2.0}
        \label{fig:comparison:hmr}
    \end{subfigure}

    \caption{\textbf{Visual comparison of 3D human mesh estimation results.} AionHMR, clearly, generates the meshes with the most accurate children's shape, while the pose and the adult reconstruction are also accurate.}
    \label{fig:comparison}
    \vspace{-.5cm}
\end{figure*}

\subsubsection{Quantitative Evaluation}

\paragraph{Height}
Table~\ref{tab:evaluation_results_single} shows the general 3D evaluation results across all datasets using the AHD and APHD metrics. AionHMR demonstrates very good performance, particularly when considering subjects with non-adult anthropometry. The overall lowest AHD and APHD confirm the superiority of our approach in handling diverse human scales, a result most clearly highlighted by the SyRIP dataset, which is dedicated exclusively to infants and toddlers. Here, AionHMR achieves a significantly lower APHD than all competing methods, confirming the effectiveness of incorporating the $\text{SMPL-A}$ model and specialized training strategies for accurately estimating highly non-adult body shapes.

In Table~\ref{tab:avg_height}, we present the average predicted height (in meters) from AionHMR and AionHMR-a compared to the evaluation baselines. This specific analysis is conducted only on the SyRIP and BabyRobot datasets because these are the only two evaluation sets comprised exclusively of infants and children, respectively.

\begin{table}

    \centering
    \caption{\textbf{Average predicted height} (in meters) comparison across different methods. SyRIP dataset contains infants only images, while BabyRobot images with children 6-10 years old. AionHMR and AionHMR-A estimate the most plausible height values on SyRIP, as well as the closest average height to the ground truth value provided by BabyRobot.}
    \label{tab:avg_height}
    \scalebox{0.9}{
    \begin{tabular}{l c c}
        \toprule
        \textbf{Method} & \textbf{SyRIP} & \textbf{BabyRobot} \\
        \midrule
        TokenHMR & 1.696 & 1.738 \\
        SLAHMR & 1.710 & 1.701 \\
        HMR2.0 & 1.705 & 1.717 \\
        BEV & 1.249 & 1.528 \\
        \midrule
        \textbf{AionHMR-a} & \textbf{0.600} & 1.250 \\
        \textbf{AionHMR} & \textbf{0.700} & \textbf{1.460} \\
        \midrule
        Ground-Truth Height & \multicolumn{1}{c}{---} & 1.330 \\
        \bottomrule
    \end{tabular}
    }
        
    \vspace{-.5cm}
\end{table}

Based on the results, the superiority of our framework is evident. The SyRIP dataset, comprised only of infants, shows that baseline methods significantly overestimate average predicted heights, often reporting highly unrealistic values due to reliance on adult geometric priors. In stark contrast, the specialized optimization of AionHMR-a yields an average height of 0.6m, a plausible number for infants. Similarly, AionHMR, trained with specialized data and the $\text{SMPL-A}$ model, achieves 0.7m, providing a more realistic and biologically plausible result. For the BabyRobot dataset, where RGB-D data enabled ground-truth calculation, both AionHMR-a and AionHMR achieve the most accurate average height estimation, further validating the necessity of a child-specific body modeling strategy.

\vspace{-0.6cm}
\paragraph{3D Pose}

\begin{table}[h!]
\centering
\caption{\textbf{3D Pose Estimation}. Model Evaluation using MPJPE (in mm). Best results are shown in \underline{underline}, while best results for the age-inclusive models are shown in \textbf{bold}. AionHMR outperforms BEV, the model with the best 3D child shape estimation and is competitive with, or even superior to, TokenHMR and HMR2.0.}
\label{tab:evaluation_results_mpjpe}
\scalebox{0.75}{ 
\begin{tabular}{l c c c c}
\toprule
\textbf{Method} & \textbf{Ages} & \textbf{SyRIP} & \textbf{ChildPlay} & \textbf{BabyRobot}\\
\midrule
TokenHMR~\cite{dwivedi_cvpr2024_tokenhmr} & Adult & 126.66 & 489.20 & 280.09 \\
HMR2.0~\cite{goel2023humans} & Adult & \underline{55.47} & \underline{236.87} & \underline{252.08}\\
\midrule
BEV~\cite{BEV} & Age-Inclusive & 452.73 & 424.41 & 380.00 \\
\midrule
\textbf{AionHMR} & Age-Inclusive & \textbf{255.01} & \textbf{283.08} & \textbf{259.92}\\
\bottomrule
\end{tabular}
}

\vspace{-0.3cm}
\end{table}
For 3D pose accuracy, we present the evaluation results in Table~\ref{tab:evaluation_results_mpjpe} using the primary metric, MPJPE, comparing AionHMR with the baselines across the three key datasets. The evaluation of the 3D pose shows that AionHMR surpasses BEV, the method that can estimate the 3D shape of children and infants more accurately, in all datasets. 
AionHMR achieves almost the same MPJPE with HMR2.0 in BabyRobot, and surpasses TokenHMR in ChildPlay and BabyRobot, showing its effectiveness in 3D pose estimation.

\begin{table*}[t!]
\centering
\caption{
\textbf{2D Pose Evaluation}. PCK scores of projected keypoints at different thresholds (higher is better). Best results per metric and dataset are shown in \underline{underline}, while best results for the age-inclusive models are shown in \textbf{bold}. Among age-inclusive models, AionHMR provides the most accurate 2D pose estimation in most datasets, and performs competitively or better than adult methods. 
}
\label{tab:evaluation_results_pck_new}
\setlength{\tabcolsep}{4pt} 
\renewcommand{\arraystretch}{1.1} 
\resizebox{\textwidth}{!}{ 
\small
\scalebox{0.9}{
\begin{tabular}{l c c c c c c c c c c c}
\toprule
\textbf{Dataset} & \multicolumn{1}{c}{} & \multicolumn{2}{c}{\textbf{SyRIP}} & \multicolumn{2}{c}{\textbf{Relative Human}} & \multicolumn{2}{c}{\textbf{ChildPlay}} & \multicolumn{2}{c}{\textbf{BabyRobot}} & \multicolumn{2}{c}{\textbf{COCO}} \\
\cmidrule(lr){1-1} \cmidrule(lr){2-2} \cmidrule(lr){3-4} \cmidrule(lr){5-6} \cmidrule(lr){7-8} \cmidrule(lr){9-10} \cmidrule(lr){11-12}
\textbf{Method} & \multicolumn{1}{c}{\textbf{Ages}} & PCK @0.05 & PCK @0.1 & PCK @0.05 & PCK @0.1 & PCK @0.05 & PCK @0.1 & PCK @0.05 & PCK @0.1 & PCK @0.05 & PCK @0.1 \\
\midrule
TokenHMR~\cite{dwivedi_cvpr2024_tokenhmr} & Adults & 0.53 & 0.90 & 0.30 & 0.45 & 0.81 & 0.89 & 0.90 & 0.99 & 0.77 & 0.95\\
HMR2.0~\cite{goel2023humans} & Adults & \underline{0.79} & \underline{0.98} & \underline{0.48} & \underline{0.62} & \underline{0.76} & \underline{0.94} & \underline{0.97} & \underline{0.99} & \underline{0.86} & \underline{0.97}\\
\midrule
BEV~\cite{BEV} & Age-Inclusive & 0.34 & 0.57 & \textbf{0.32} & \textbf{0.55} & 0.43 & 0.73 & 0.61 & 0.86 & 0.47& 0.66\\
\midrule
\textbf{AionHMR} & Age-Inclusive & \textbf{0.67} & \textbf{0.91} & 0.30 & 0.44 & \textbf{0.58} & \textbf{0.84} & \textbf{0.93} & \textbf{0.98} & \textbf{0.79} & \textbf{0.93} \\
\bottomrule
\end{tabular}
}
} 

\vspace{-0.5cm}
\end{table*}

\vspace{-0.5cm}
\paragraph{2D Pose} Finally, in Table~\ref{tab:evaluation_results_pck_new}, we present the results of the 2D pose evaluation using the PCK metric with two different thresholds (0.05 and 0.1). For this comparison, all available evaluation datasets are used to assess the robustness of the methods' 2D keypoint prediction. Based on the results, AionHMR effectively estimates 2D pose across all datasets, outperforming the BEV method, which is designed to model children, on every dataset except Relative Human. The superior performance of BEV on the Relative Human dataset is attributed to its inclusion of this dataset in its training. The HMR2.0 model consistently achieves the highest PCK score for 2D pose estimation across all datasets. While AionHMR was trained on the same datasets as HMR2.0, the difference in results can be attributed to architectural variations and the incorporation of the SMPL-A body model, in contrast to HMR2.0's use of the standard SMPL model. Finally, our competitive results on COCO prove the efficacy of AionHMR in adult mesh recovery.

\subsubsection{Qualitative Results - Subjective Study}
Qualitative evaluation remains an essential component of 3D vision research, as certain perceptual aspects, such as the plausibility of body proportions, the absence of self-intersections, and the smoothness of reconstructed surfaces, are not fully captured by quantitative metrics. Therefore, visual inspection provides valuable insight into model performance. As illustrated in Figures~\ref{fig:teaser} and~\ref{fig:BRdataset}, and through the comparisons with HMR2.0 and BEV shown in Figure~\ref{fig:comparison}, AionHMR produces more realistic and consistent reconstructions, highlighting its qualitative advantages over existing HMR approaches.

\vspace{-0.5cm}
\paragraph{Setup}To quantitatively assess perceptual quality, we conducted a comprehensive, pairwise, subjective study with 30 participants, comparing AionHMR with HMR2.0 and BEV. We utilized a test set of 39 images, sampled proportionally from our evaluation datasets, and generated 3D reconstructions from all three methods. Each participant completed 25 randomized paired comparisons, selecting the superior reconstruction based on three criteria: shape fidelity, pose accuracy and overall efficacy. The comparison structure emphasized direct competition with baselines, 20 comparisons: AionHMR vs. HMR2.0 and AionHMR vs. BEV, while 5 comparisons between HMR2.0 and BEV served as an internal benchmark.

\begin{table}[!ht]
\centering
\caption{\textbf{Subjective Study Results}: Pairwise Comparison Win Rates by Category. Participants clearly selected AionHMR as the best method in the 3 categories tested.}
\label{tab:subjective_reformatted}
\scalebox{0.72}{
\begin{tabular}{l l c c c}
\toprule
\multicolumn{2}{l}{} & \multicolumn{3}{c}{\textbf{Win Rate (\%)}}\\
\cmidrule(lr){3-5}
\textbf{Comparison Pair} & \textbf{Method} & \textbf{Shape} & \textbf{Pose} & \textbf{Overall} \\
\midrule
\multirow{3}{*}{\textbf{AionHMR vs. BEV}} & \textbf{AionHMR} & \textbf{76.33} & \textbf{76.67} & \textbf{74.67} \\
 & BEV & 16.33 & 20.00 & 15.33 \\
 & Cannot Determine & 7.34 & 3.33 & 10.00 \\
\cmidrule(lr){1-5}
\multirow{3}{*}{\textbf{AionHMR vs. HMR2.0}} & \textbf{AionHMR} & \textbf{52.33} & \textbf{49.00} & \textbf{49.33} \\
 & HMR2.0 & 29.67 & 35.33 & 32.33 \\
 & Cannot Determine & 18.00 & 15.67 & 18.34 \\
\bottomrule
\end{tabular}
}

\vspace{-0.4cm}
\end{table}

\vspace{-0.6cm}
\paragraph{Results}Based on the aggregated survey data, we derived comprehensive statistics concerning the preferred shape and pose estimation. Table~\ref{tab:subjective_reformatted} shows the results of the pairwise comparisons, AionHMR vs BEV and AionHMR vs HMR2.0.
The results of the subjective user study show a clear and consistent performance advantage for AionHMR. Our model was preferred across every category tested: shape fidelity, pose accuracy, and overall quality, when compared head-to-head with both BEV and HMR2.0.

More specifically, AionHMR demonstrated a substantial gain against one baseline, being selected in approximately 75\% of the comparison votes against BEV in all categories. While the margin was smaller against HMR2.0, AionHMR still maintained an advantage, particularly in the pose category. Aggregating the results across all categories involving AionHMR (excluding the baseline-only comparisons), AionHMR was preferred in more than 60\% of the total answers, unequivocally providing the best overall reconstructions. This establishes a clear subjective performance advantage over both baselines, with a substantial preference gap of over 10\% in the total win rate. Furthermore, the results indicate that participants perceived our shape estimation as more accurate than our pose estimation, a finding corroborated by our quantitative evaluation metrics.

\vspace{-0.3cm}

\section{Action-Preserving Data Anonymization}

\begin{figure*}[h!t]
    \centering
       \includegraphics[width=0.95\textwidth,clip]{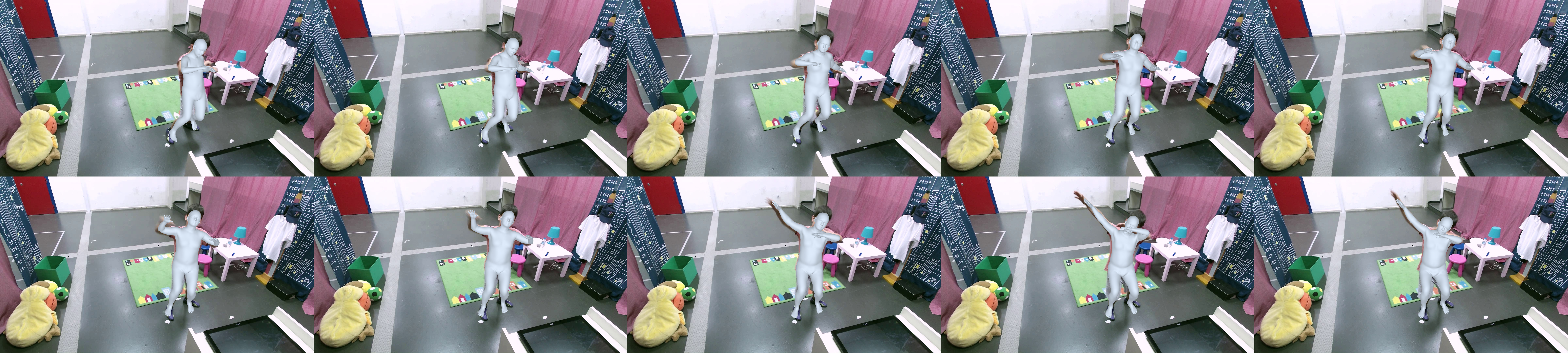}
    \caption{$\textbf{3D-BabyRobot Dataset}$. We release the 3D-BabyRobot dataset which contains 3D reconstructions of children interacting with robots. These reconstructions preserve the action and the behavior of the children and enable the release of sensitive data via anonymization.}
    \label{fig:BRdataset}
    \vspace{-0.5cm}
\end{figure*}
\vspace{-0.2cm}
Data scarcity persists in child-related computer vision research, directly attributable to the sensitive nature of these data and the stringent legal frameworks protecting minors. These necessary protections inherently restrict the availability of large-scale, unprocessed datasets, as releasing 
any human-centric data that could compromise individual identity is prohibited. Inspired by the observation that AionHMR yields accurate 3D reconstructions for humans spanning the entire age spectrum, we hypothesize that the resulting reconstructed 3D meshes can effectively serve as privacy-preserving synthetic data proxies. To test this hypothesis, we conduct a rigorous behavioral analysis and action preservation experiment.

\vspace{-0.5cm}
\paragraph{Dataset} The BabyRobot dataset consists of Child-Robot Interaction videos, a subset of which shows children executing specific actions and pantomimes for a robot. These actions involve complex movements of the arms, legs, head, and torso, providing a challenging testbed for evaluating the action, pose, and geometry preservation fidelity of the AionHMR 3D reconstructions. We perform a detailed behavioral analysis because a single action recognition experiment is insufficient to quantify subtle kinematic preservation. Therefore, we randomly select 43 video sequences covering 6 distinct actions: dancing, ``come closer'' gesture, swimming, ``draw a circle'' gesture, playing the guitar, and ``hello'' gesture.
\vspace{-0.5cm}
\paragraph{Video Descriptions} The first step is to generate the corresponding videos with the AionHMR 3D reconstructed meshes. Then, we leverage a Large Vision Language Model (LVLM)~\cite{Qwen2.5-VL} to generate a detailed description of the actions and movements in both the original video and the corresponding synthetic 3D mesh sequence. The LVLM prompt focuses on the analysis of kinematic parameters (\eg, arm, leg, and head movement, orientation) while explicitly instructing the model to disregard identifying facial or anatomical features. The core of our validation involves quantifying the semantic similarity between the paired original and synthetic descriptions.

\vspace{-0.5cm}
\paragraph{Description Similarity}To this end, we ask a Large Language Model (LLM)~\cite{comanici2025gemini} to compare the semantic similarity between the two descriptions and assign a numerical percentage similarity value. Additionally, it is important to determine the maximum and minimum values of the similarity scores from the LLM's descriptions. To determine the maximum similarity value, we generate multiple descriptions using the same prompt for the original videos and compare their similarity via the LLM. For the minimum similarity value, we create pairs of videos that do not belong to the same action and compare the similarity of their descriptions. Finally, we analyze the semantic similarity of descriptions generated from video pairs of the same action performed by different children, isolating the influence of the child's identity on the description's semantics.

\begin{table}
    \centering
\caption{\textbf{Average Similarity (\%) of the videos' descriptions} under different testing scenarios. The first three scenarios compare an original video with the video generated from AionHMR, while the last scenario compares only original videos. Descriptions from videos with AionHMR's reconstructions are 60\% similar to those of the original video, while the maximum possible value for similarity is 75.6\% (scenario 4) and the minimum 43.3\% (scenario 1).} 
         \label{tab:accuracy_scenarios_vertical}
   
    \scalebox{0.85}{
    \begin{tabular}{lc}
        \toprule
        \textbf{Scenario} & \textbf{Average Similarity (\%)} \\
        \midrule
        Different Action - Different Child & 43.3 \\
        Same Action - Different Child & 46.6 \\
        Same Action - Same Child & 60.0 \\
        Multiple Runs - Same Video & 75.6 \\
        \bottomrule
    \end{tabular}
    }
        
         \vspace{-0.5cm}
\end{table}

\vspace{-0.5cm}
\paragraph{Results}The experimental results validate that AionHMR effectively preserves human actions and behaviors. The maximum expected fidelity bound is defined at 75.6\% by the ``Multiple Runs - Same Video" scenario, which accounts for the LLM's intrinsic descriptive variability. Crucially, the ``Same Action - Same Child" scenario achieves a value less than but near enough this maximum value, indicating that the AionHMR reconstruction retains the original action's kinematic features to the highest possible degree. 
The ``Same Action - Different Child" result (46.6\%) highlights inter-subject variability. This lower similarity demonstrates the descriptions' sensitivity to the unique kinematic style of the individual child. This suggests that the AionHMR reconstruction successfully preserves the subject-specific behavioral signature, not merely the action class, due to the accurate 3D representations it produces.

\vspace{-0.55cm}
\paragraph{3D-BabyRobot Dataset} These behavioral analysis results, coupled with the accurate 3D reconstructions produced by AionHMR, enable us to release the \textbf{3D-BabyRobot Dataset}. This dataset consists of the privacy-preserving 3D mesh reconstructions of children interacting with robots, which, as demonstrated, successfully retain the essential pose, behavior, action, and geometry information about the subjects. This practice exemplifies effective sensitive data anonymization, thereby facilitating the release of valuable large-scale data for child-centric computer vision research. An example from the 3D-BabyRobot dataset is shown in Figure~\ref{fig:BRdataset}, depicting video frames of a child interacting with a robot and the corresponding 3D reconstructions. We release approximately 135k frames with 3D reconstructions per child for 28 children, totaling over 4 million annotated frames, as shown in Table~\ref{tab:annotated}. The table also reports the number of annotated training samples we provide from SyRIP and Relative Human.

\vspace{-0.2cm}
\begin{table}[!h]
  \centering
  \caption{\textbf{Number of annotated samples per dataset in the released collection.} We provide large-scale datasets with high-quality annotations for the training of HMR-like models or other Computer Vision applications.}  
    \label{tab:annotated}
    \scalebox{0.9}{
    \begin{tabular}{l c c c}
    \toprule
      \textbf{Dataset}   & SyRIP & Relative Human & 3D-BabyRobot\\
      \midrule
    \textbf{\# Samples} & 509 & 1052 & 4M \\
    \bottomrule
    \end{tabular}
    }
    \vspace{-.2cm}
    
    \vspace{-.5cm}
\end{table}

\section{Conclusion}
\label{sec:conclusion}

In this work, we propose AionHMR, a unified framework that contributes to bridging the current domain gap in 3D human shape and pose estimation from images, when the input is infants or children. The two parts of AionHMR, the high-quality optimization-based method (AionHMR-a) and the trained transformer-based network (AionHMR-b), quantitatively and qualitatively demonstrate state-of-the-art performance, establishing a new benchmark against modern, similar works in the challenging domain of non-adult 3D shape estimation. Moreover, we establish an effective methodology for action-preserving ethically releasing sensitive data by sharing 3D human reconstructions instead of raw imagery. This process inherently anonymizes the identity of the subjects (children and infants) while providing accurate 3D body and motion information essential for subsequent motion analysis and action recognition tasks. The 3D-BabyRobot dataset exemplifies this approach, offering 3D reconstructions of children interacting with robots, complete with diverse actions, gestures, and spatial movements.
\vspace{-0.7cm}
\paragraph{Limitations and Future Work} 
While AionHMR advances Human Mesh Recovery for non-adult populations, several challenges remain. Incorporating multi-view or video data could provide a more accurate estimation of body pose and shape, especially in in-the-wild scenarios with complex motions. Moreover, the SMPL-A model inherits shape-space biases from SMPL, limiting representation fidelity for children and infants. Future work will explore a new unified all-age body model capturing the full spectrum of human morphology for more accurate and generalizable reconstructions.

\section{Acknowledgments}
This work was partially funded by the European Union under Horizon Europe (grant No. 101136568 - HERON).
\vspace{-0.45cm}
\begin{figure}[H]
    \includegraphics[width=0.5\linewidth]{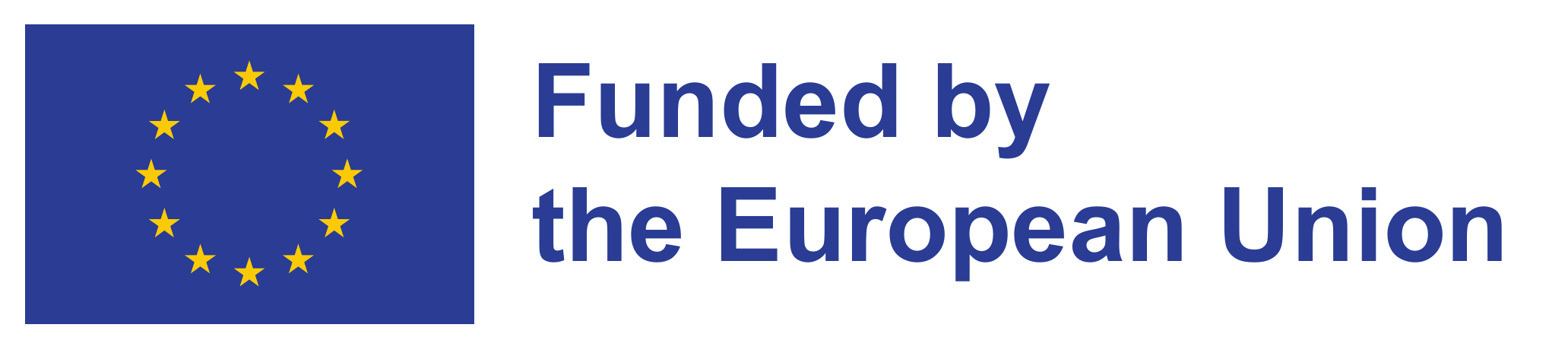}
\end{figure}

{
    \small
    \bibliographystyle{ieeenat_fullname}
    \bibliography{main}

@String(PAMI = {IEEE Trans. Pattern Anal. Mach. Intell.})

@String(CVPR= {IEEE Conf. Comput. Vis. Pattern Recog.})

@String(ICCV= {Int. Conf. Comput. Vis.})

@String(ECCV= {Eur. Conf. Comput. Vis.})

@String(TOG= {ACM Trans. Graph.})

@String(ICME = {Int. Conf. Multimedia and Expo})

@String(ICLR = {Int. Conf. Learn. Represent.})

@String(PAMI  = {IEEE TPAMI})

@String(CVPR  = {CVPR})

@String(ICCV  = {ICCV})

@String(ECCV  = {ECCV})

@String(TOG   = {ACM TOG})

@String(ICME  =	{ICME})

@String(ICLR  = {ICLR})

@inproceedings{dosovitskiy2021imageworth16x16words,
	title        = {An Image is Worth 16x16 Words: Transformers for Image Recognition at Scale},
	author       = {Alexey Dosovitskiy and Lucas Beyer and Alexander Kolesnikov and Dirk Weissenborn and Xiaohua Zhai and Thomas Unterthiner and Mostafa Dehghani and Matthias Minderer and Georg Heigold and Sylvain Gelly and Jakob Uszkoreit and Neil Houlsby},
	year         = 2021,
	booktitle    = {International Conference on Learning Representations (ICLR)}
}

@article{efthymiou2022childbot,
  title={ChildBot: Multi-robot perception and interaction with children},
  author={Efthymiou, Niki and Filntisis, Panagiotis P and Koutras, Petros and Tsiami, Antigoni and Hadfield, Jack and Potamianos, Gerasimos and Maragos, Petros},
  journal={Robotics and Autonomous Systems},
  volume={150},
  pages={103975},
  year={2022},
  publisher={Elsevier}
}

@article{SMPL:2015,
	title        = {{SMPL}: A Skinned Multi-Person Linear Model},
	author       = {Loper, Matthew and Mahmood, Naureen and Romero, Javier and Pons-Moll, Gerard and Black, Michael J.},
	year         = 2015,
	journal      = {ACM Trans. Graphics (Proc. SIGGRAPH Asia)},
      pages = {248:1--248:16},
      volume = {34},
}

@inproceedings{Bogo:ECCV:2016,
	title        = {Keep it {SMPL}: Automatic Estimation of {3D} Human Pose and Shape from a Single Image},
	author       = {Bogo, Federica and Kanazawa, Angjoo and Lassner, Christoph and Gehler, Peter and Romero, Javier and Black, Michael J.},
	year         = 2016,
	booktitle    = {European Conference on Computer Vision (ECCV)},
}

@inproceedings{SMPL-X:2019,
	title        = {Expressive Body Capture: 3D Hands, Face, and Body from a Single Image},
	author       = {Pavlakos, Georgios and Choutas, Vasileios and Ghorbani, Nima and Bolkart, Timo and Osman, Ahmed AA and Tzionas, Dimitrios and Black, Michael J},
	year         = 2019,
	booktitle    = {IEEE/CVF Conference on Computer Vision and Pattern Recognition (CVPR)}
}

@inproceedings{ye2023slahmr,
	title        = {Decoupling Human and Camera Motion from Videos in the Wild},
	author       = {Ye, Vickie and Pavlakos, Georgios and Malik, Jitendra and Kanazawa, Angjoo},
	year         = 2023,
	booktitle    = {IEEE/CVF Conference on Computer Vision and Pattern Recognition (CVPR)}
}

@inproceedings{goel2023humans,
	title        = {Humans in 4{D}: Reconstructing and Tracking Humans with Transformers},
	author       = {Goel, Shubham and Pavlakos, Georgios and Rajasegaran, Jathushan and Kanazawa, Angjoo and Malik, Jitendra},
	year         = 2023,
	booktitle    = {International Conference on Computer Vision (ICCV)}
}

@inproceedings{hmrKanazawa17,
	title        = {End-to-end Recovery of Human Shape and Pose},
	author       = {Angjoo Kanazawa and Michael J. Black and David W. Jacobs and Jitendra Malik},
	year         = 2018,
	booktitle    = {IEEE/CVF Conference on Computer Vision and Pattern Recognition (CVPR)}
}

@inproceedings{NIPS2017_3f5ee243,
	title        = {Attention is All you Need},
	author       = {Vaswani, Ashish and Shazeer, Noam and Parmar, Niki and Uszkoreit, Jakob and Jones, Llion and Gomez, Aidan N and Kaiser, {\L}ukasz and Polosukhin, Illia},
	year         = 2017,
	booktitle    = {Advances in Neural Information Processing Systems (NeurIPS)},

}

@article{teed2021droid,
	title        = {{DROID-SLAM: Deep Visual SLAM for Monocular, Stereo, and RGB-D Cameras}},
	author       = {Teed, Zachary and Deng, Jia},
	year         = 2021,
	journal      = {Advances in Neural Information Processing Systems (NeurIPS)}
}

@inproceedings{rajasegaran2022tracking,
	title        = {Tracking People by Predicting 3D Appearance, Location and Pose},
	author       = {Rajasegaran, Jathushan and Pavlakos, Georgios and Kanazawa, Angjoo and Malik, Jitendra},
	year         = 2022,
	booktitle    = {IEEE/CVF Conference on Computer Vision and Pattern Recognition (CVPR)},
}

@inproceedings{Patel:CVPR:2021,
	title        = {{AGORA}: Avatars in Geography Optimized for Regression Analysis},
	author       = {Patel, Priyanka and Huang, Chun-Hao P. and Tesch, Joachim and Hoffmann, David T. and Tripathi, Shashank and Black, Michael J.},
	year         = 2021,
	booktitle    = {IEEE/CVF Conference on Computer Vision and Pattern Recognition (CVPR)},

}

@inproceedings{10.1007/978-3-030-00928-1_89,
	title        = {Learning an Infant Body Model from RGB-D Data for Accurate Full Body Motion Analysis},
	author       = {Hesse, Nikolas and Pujades, Sergi and Romero, Javier and Black, Michael J. and Bodensteiner, Christoph and Arens, Michael and Hofmann, Ulrich G. and Tacke, Uta and Hadders-Algra, Mijna and Weinberger, Raphael and M\"{u}ller-Felber, Wolfgang and Sebastian Schroeder, A.},
	year         = 2018,
	booktitle    =  {Medical Image Computing and Computer Assisted Intervention (MICCAI)},
}

@article{MANO:SIGGRAPHASIA:2017,
	title        = {Embodied Hands: Modeling and Capturing Hands and Bodies Together},
	author       = {Romero, Javier and Tzionas, Dimitrios and Black, Michael J.},
	year         = 2017,
	journal      = {ACM Transactions on Graphics (ToG)},
 volume = {36},
      series = {245:1--245:17},
}

@inproceedings{xu2022vitpose,
	title        = {Vi{TP}ose: Simple Vision Transformer Baselines for Human Pose Estimation},
	author       = {Yufei Xu and Jing Zhang and Qiming Zhang and Dacheng Tao},
	year         = 2022,
	booktitle    = {Advances in Neural Information Processing Systems (NeurIPS)}
}

@inproceedings{barron2019generaladaptiverobustloss,
	title        = {A General and Adaptive Robust Loss Function},
	author       = {Jonathan T. Barron},
	year         = 2019,
	  booktitle    = {IEEE/CVF Conference on Computer Vision and Pattern Recognition (CVPR)}, 
 
}

@inproceedings{xia2025reconstructinghumansbiomechanicallyaccurate,
	title        = {Reconstructing Humans with a Biomechanically Accurate Skeleton},
	author       = {Yan Xia and Xiaowei Zhou and Etienne Vouga and Qixing Huang and Georgios Pavlakos},
	year         = 2025,
	booktitle    = {IEEE/CVF Conference on Computer Vision and Pattern Recognition (CVPR)}
}

@inproceedings{koleini2025bioposebiomechanicallyaccurate3dpose,
	title        = {BioPose: Biomechanically-accurate 3D Pose Estimation from Monocular Videos},
	author       = {Farnoosh Koleini and Muhammad Usama Saleem and Pu Wang and Hongfei Xue and Ahmed Helmy and Abbey Fenwick},
	year         = 2025,
	booktitle    = {IEEE/CVF Winter Conference on Applications of Computer Vision (WACV)}
}

@inproceedings{chen20173dhumanposeestimation,
	title        = {3D Human Pose Estimation = 2D Pose Estimation + Matching},
	author       = {Ching-Hang Chen and Deva Ramanan},
	year         = 2017,
	booktitle    = {IEEE/CVF Conference on Computer Vision and Pattern Recognition (CVPR)}
}

@article{Hofmann2012,
	title        = {Multi-view 3D Human Pose Estimation in Complex Environment},
	author       = {Hofmann, M. and Gavrila, D. M.},
	year         = 2012,
	journal={International Journal of Computer Vision},
  volume={96},
  pages={103--124},
}

@inproceedings{dong2021shapeawaremultipersonposeestimation,
	title        = {Shape-aware Multi-Person Pose Estimation from Multi-View Images},
	author       = {Zijian Dong and Jie Song and Xu Chen and Chen Guo and Otmar Hilliges},
	year         = 2021,
	booktitle    = {International Conference on Computer Vision (ICCV)}
}

@inproceedings{pavlakos2017harvestingmultipleviewsmarkerless,
	title        = {Harvesting Multiple Views for Marker-less 3D Human Pose Annotations},
	author       = {Georgios Pavlakos and Xiaowei Zhou and Konstantinos G. Derpanis and Kostas Daniilidis},
	year         = 2017,
	booktitle    = {IEEE/CVF Conference on Computer Vision and Pattern Recognition (CVPR)}
}

@inproceedings{martinez2017simple,
	title        = {A simple yet effective baseline for 3d human pose estimation},
	author       = {Martinez, Julieta and Hossain, Rayat and Romero, Javier and Little, James J},
	year         = 2017,
	booktitle    = {International Conference on Computer Vision (ICCV)},
}

@inproceedings{moreno20173d,
	title        = {3d human pose estimation from a single image via distance matrix regression},
	author       = {Moreno-Noguer, Francesc},
	year         = 2017,
	booktitle    = {IEEE/CVF Conference on Computer Vision and Pattern Recognition (CVPR)},
}

@inproceedings{pavlakos2022multishot,
	title        = {Human Mesh Recovery from Multiple Shots},
	author       = {Pavlakos, Georgios and Malik, Jitendra and Kanazawa, Angjoo},
	year         = 2022,
	booktitle    = {IEEE/CVF Conference on Computer Vision and Pattern Recognition (CVPR)}
}

@inproceedings{park20163dhumanposeestimation,
	title        = {3D Human Pose Estimation Using Convolutional Neural Networks with 2D Pose Information},
	author       = {Sungheon Park and Jihye Hwang and Nojun Kwak},
	year         = 2016,
	booktitle    = {European Conference on Computer Vision (ECCV)}
}

@inproceedings{person3dyan,
	title        = {Person-in-WiFi 3D: End-to-End Multi-Person 3D Pose Estimation with Wi-Fi},
	author       = {Yan, Kangwei and Wang, Fei and Qian, Bo and Ding, Han and Han, Jinsong and Wei, Xing},
	year         = 2024,
	booktitle    = {IEEE/CVF Conference on Computer Vision and Pattern Recognition (CVPR)}
}

@inproceedings{liao2023multipleviewgeometrytransformers,
	title        = {Multiple View Geometry Transformers for 3D Human Pose Estimation},
	author       = {Ziwei Liao and Jialiang Zhu and Chunyu Wang and Han Hu and Steven L. Waslander},
	year         = 2023,
	booktitle    = {IEEE/CVF Conference on Computer Vision and Pattern Recognition (CVPR)}
}

@inproceedings{motionagformer2024,
	title        = {MotionAGFormer: Enhancing 3D Human Pose Estimation with a Transformer-GCNFormer Network},
	author       = {Soroush Mehraban, Vida Adeli, Babak Taati},
	year         = 2024,
	booktitle    = {IEEE/CVF Winter Conference on Applications of Computer Vision (WACV)}
}

@inproceedings{bashirov2021realtimergbdbasedextendedbody,
	title        = {Real-time RGBD-based Extended Body Pose Estimation},
	author       = {Bashirov, Renat and Ianina, Anastasia and Iskakov, Karim and Kononenko, Yevgeniy and Strizhkova, Valeriya and Lempitsky, Victor and Vakhitov, Alexander},
	year         = 2021,
	booktitle    = {IEEE/CVF Winter Conference on Applications of Computer Vision (WACV)}
}

@inproceedings{fan2021revitalizingoptimization3dhuman,
	title        = {Revitalizing Optimization for 3D Human Pose and Shape Estimation: A Sparse Constrained Formulation},
	author       = {Taosha Fan and Kalyan Vasudev Alwala and Donglai Xiang and Weipeng Xu and Todd Murphey and Mustafa Mukadam},
	year         = 2021,
	booktitle    = {International Conference on Computer Vision (ICCV)}
}

@inproceedings{Zheng2019DeepHuman,
	title        = {DeepHuman: 3D Human Reconstruction from a Single Image},
	author       = {Zheng, Zerong and Yu, Tao and Wei, Yixuan and Dai, Qionghai and Liu, Yebin},
	year         = 2019,
	booktitle    = {International Conference on Computer Vision (ICCV)}
}

@inproceedings{kolotouros2019spin,
	title        = {Learning to Reconstruct 3D Human Pose and Shape via Model-fitting in the Loop},
	author       = {Kolotouros, Nikos and Pavlakos, Georgios and Black, Michael J and Daniilidis, Kostas},
	year         = 2019,
	booktitle    = {International Conference on Computer Vision (ICCV)}
}

@inproceedings{BEV,
	title        = {Putting People in their Place: Monocular Regression of {3D} People in Depth},
	author       = {Sun, Yu and Liu, Wu and Bao, Qian and Fu, Yili and Mei, Tao and Black, Michael J.},
	year         = 2022,
	booktitle    = {IEEE/CVF Conference on Computer Vision and Pattern Recognition (CVPR)}
}

@inproceedings{huang2021infant,
	title        = {Invariant Representation Learning for Infant Pose Estimation with Small Data},
	author       = {Huang, Xiaofei and Fu, Nihang and Liu, Shuangjun and Ostadabbas, Sarah},
	year         = 2021,
	booktitle    = {IEEE International Conference on Automatic Face and Gesture Recognition}
}

@article{ionescu2013human3,
	title        = {Human3. 6m: Large scale datasets and predictive methods for 3d human sensing in natural environments},
	author       = {Ionescu, Catalin and Papava, Dragos and Olaru, Vlad and Sminchisescu, Cristian},
	year         = 2013,
	journal      = {IEEE Transactions on Pattern Analysis and Machine Intelligence (PAMI)},
   volume={36},
  pages={1325--1339},

}

@inproceedings{mehta2017monocular,
	title        = {Monocular 3d human pose estimation in the wild using improved cnn supervision},
	author       = {Mehta, Dushyant and Rhodin, Helge and Casas, Dan and Fua, Pascal and Sotnychenko, Oleksandr and Xu, Weipeng and Theobalt, Christian},
	year         = 2017,
	booktitle    = {International Conference on 3D Vision (3DV)},
}

@inproceedings{lin2014microsoft,
	title        = {Microsoft coco: Common objects in context},
	author       = {Tsung-Yi Lin and Michael Maire and Serge Belongie and Lubomir Bourdev and Ross Girshick and James Hays and Pietro Perona and Deva Ramanan and C. Lawrence Zitnick and Piotr Dollár},
	year         = 2014,
	booktitle    = {European Conference on Computer Vision (ECCV)},
}

@inproceedings{andriluka20142d,
	title        = {2d human pose estimation: New benchmark and state of the art analysis},
	author       = {Andriluka, Mykhaylo and Pishchulin, Leonid and Gehler, Peter and Schiele, Bernt},
	year         = 2014,
	booktitle    = {IEEE/CVF Conference on Computer Vision and Pattern Recognition (CVPR)},
}

@inproceedings{kolotouros2021probabilistic,
	title        = {Probabilistic modeling for human mesh recovery},
	author       = {Kolotouros, Nikos and Pavlakos, Georgios and Jayaraman, Dinesh and Daniilidis, Kostas},
	year         = 2021,
	booktitle    = {International Conference on Computer Vision (ICCV)},
}

@article{kojovic2021using,
	title        = {Using 2D video-based pose estimation for automated prediction of autism spectrum disorders in young children},
	author       = {Kojovic, Nada and Natraj, Shreyasvi and Mohanty, Sharada Prasanna and Maillart, Thomas and Schaer, Marie},
	year         = 2021,
	journal      = {Scientific Reports},
 volume={11},
  pages={15069},
}

@article{ganai2025early,
	title        = {Early detection of autism spectrum disorder: gait deviations and machine learning},
	author       = {Ganai, Umer Jon and Ratne, Aditya and Bhushan, Braj and Venkatesh, KS},
	year         = 2025,
	journal      = {Scientific Reports},
  volume={15},
  pages={873},
}

@inproceedings{10.5555/3454287.3455008,
	title        = {PyTorch: an imperative style, high-performance deep learning library},
	author       = {Paszke, Adam and Gross, Sam and Massa, Francisco and Lerer, Adam and Bradbury, James and Chanan, Gregory and Killeen, Trevor and Lin, Zeming and Gimelshein, Natalia and Antiga, Luca and Desmaison, Alban and K\"{o}pf, Andreas and Yang, Edward and DeVito, Zach and Raison, Martin and Tejani, Alykhan and Chilamkurthy, Sasank and Steiner, Benoit and Fang, Lu and Bai, Junjie and Chintala, Soumith},
	year         = 2019,
	booktitle    = {Advances in Neural Information Processing Systems (NeurIPS)},
}

@inproceedings{loshchilov2017decoupled,
	title        = {Decoupled weight decay regularization},
	author       = {Loshchilov, Ilya and Hutter, Frank},
	year         = 2017,
	booktitle      = {International Conference on Learning Representations (ICLR)}
}

@inproceedings{kanazawa2019learning,
	title        = {Learning 3d human dynamics from video},
	author       = {Kanazawa, Angjoo and Zhang, Jason Y and Felsen, Panna and Malik, Jitendra},
	year         = 2019,
	booktitle    = {IEEE/CVF Conference on Computer Vision and Pattern Recognition (CVPR)},
}

@inproceedings{gu2018ava,
	title        = {Ava: A video dataset of spatio-temporally localized atomic visual actions},
	author       = {Chunhui Gu and Chen Sun and David A. Ross and Carl Vondrick and Caroline Pantofaru and Yeqing Li and Sudheendra Vijayanarasimhan and George Toderici and Susanna Ricco and Rahul Sukthankar and Cordelia Schmid and Jitendra Malik},
	year         = 2018,
	booktitle    = {IEEE/CVF Conference on Computer Vision and Pattern Recognition (CVPR)},
}

@inproceedings{wu2017ai,
	title        = {Large-Scale Datasets for Going Deeper in Image Understanding},
	author       = {Wu, Jiahong and Zheng, He and Zhao, Bo and Li, Yixin and Yan, Baoming and Liang, Rui and Wang, Wenjia and Zhou, Shipei and Lin, Guosen and Fu, Yanwei and Wang, Yizhou and Wang, Yonggang},
	year         = 2019,
	booktitle      = {IEEE International Conference on Multimedia and Expo (ICME)}
}

@inbook{Roussos2025three,
  author    = {Roussos, A. and Maragos, P.},
  title     = {Three-Dimensional Modeling of Deformable Objects},
  booktitle = {Topics in Computer Vision and Machine Learning},
  publisher = {Kallipos, Open Academic Editions},
  year      = {2025},
  note      = {[Chapter]},
  url       = {https://hdl.handle.net/11419/15135}
}

@conference{dwivedi_cvpr2024_tokenhmr,
  title = {{TokenHMR}: Advancing Human Mesh Recovery with a Tokenized Pose Representation},
  author = {Dwivedi, Sai Kumar and Sun, Yu and Patel, Priyanka and Feng, Yao and Black, Michael J.},
  booktitle =  {IEEE/CVF Conference on Computer Vision and Pattern Recognition (CVPR)},
  year = {2024},
}

@inproceedings{belghit2018vision,
  title={Vision-based Pose Estimation for Augmented Reality: Comparison Study},
  author={Belghit, Hayet and Bellarbi, Abdelkader and Zenati, Nadia and Otmane, Samir},
  booktitle={3rd IEEE International Conference on Pattern Analysis and Intelligent Systems (PAIS 2018)},
  year={2018},
  organization={IEEE}
}

@inproceedings{baumgartner2023monocular,
  title={Monocular 3d human pose estimation for sports broadcasts using partial sports field registration},
  author={Baumgartner, Tobias and Klatt, Stefanie},
  booktitle={Proceedings of the IEEE/CVF conference on computer vision and pattern recognition},
  year={2023}
}

@article{Qwen2.5-VL,
  title={Qwen2.5-VL Technical Report},
  author={Bai, Shuai and Chen, Keqin and Liu, Xuejing and Wang, Jialin and Ge, Wenbin and Song, Sibo and Dang, Kai and Wang, Peng and Wang, Shijie and Tang, Jun and Zhong, Humen and Zhu, Yuanzhi and Yang, Mingkun and Li, Zhaohai and Wan, Jianqiang and Wang, Pengfei and Ding, Wei and Fu, Zheren and Xu, Yiheng and Ye, Jiabo and Zhang, Xi and Xie, Tianbao and Cheng, Zesen and Zhang, Hang and Yang, Zhibo and Xu, Haiyang and Lin, Junyang},
  journal={arXiv preprint arXiv:2502.13923},
  year={2025}
}

@article{comanici2025gemini,
  title={Gemini 2.5: Pushing the frontier with advanced reasoning, multimodality, long context, and next generation agentic capabilities},
  author={Comanici, Gheorghe and Bieber, Eric and Schaekermann, Mike and Pasupat, Ice and Sachdeva, Noveen and Dhillon, Inderjit and Blistein, Marcel and Ram, Ori and Zhang, Dan and Rosen, Evan and others},
  journal={arXiv preprint arXiv:2507.06261},
  year={2025}
}

@inproceedings{rajasegaran2023benefits,
  title={On the benefits of 3d pose and tracking for human action recognition},
  author={Rajasegaran, Jathushan and Pavlakos, Georgios and Kanazawa, Angjoo and Feichtenhofer, Christoph and Malik, Jitendra},
  booktitle={Proceedings of the IEEE/CVF conference on computer vision and pattern recognition},
  year={2023}
}

@INPROCEEDINGS{9881695,
  author={Martini, Enrico and Boldo, Michele and Aldegheri, Stefano and De Marchi, Mirco and Valè, Nicola and Filippetti, Mirko and Smania, Nicola and Bertucco, Matteo and Picelli, Alessandro and Bombieri, Nicola},
  booktitle={2022 18th International Conference on Distributed Computing in Sensor Systems (DCOSS)}, 
  title={Real-time Human Pose Estimation at the Edge for Gait Analysis at a Distance}, 
  year={2022},
}

@inproceedings{10.1007/978-981-99-6498-7_16,
author = {Huo, Rongtian and Gao, Qing and Qi, Jing and Ju, Zhaojie},
title = {3D Human Pose Estimation in Video for Human-Computer/Robot Interaction},
year = {2023},

booktitle = {Intelligent Robotics and Applications ICIRA 2023},
}

@inproceedings{tafasca2023childplay,
  title={Childplay: A new benchmark for understanding children's gaze behaviour},
  author={Tafasca, Samy and Gupta, Anshul and Odobez, Jean-Marc},
  booktitle={Proceedings of the IEEE/CVF International Conference on Computer Vision},
  year={2023}
}
}

\appendix
\clearpage
\setcounter{page}{1}
\maketitlesupplementary

This supplementary material provides additional technical details and results concerning our proposed method, AionHMR, and the newly introduced 3D-BabyRobot Dataset. The document is organized into four main sections: Section~\ref{sec:aiona} details the optimization pipeline used for the AionHMR-a; Section~\ref{sec:aionb} provides a deeper technical dive into AionHMR-b, specifically outlining its architectural configuration, a comprehensive breakdown of all utilized loss functions, and the composition of its training datasets; Section~\ref{sec:action} describes the Action Preservation Test conducted; and finally, Section~\ref{sec:qual} presents extensive qualitative results, including a large number of example images from the 3D-BabyRobot Dataset, along with visual comparisons illustrating the performance of AionHMR against two baselines: HMR2.0~\cite{goel2023humans} and BEV~\cite{BEV}.

\section{AionHMR-a Details}
\label{sec:aiona}

\subsection{Optimization Stages}
For the optimization phase of AionHMR-a, we adapt parts of the optimization process from SLAHMR~\cite{ye2023slahmr}, which we present more formally here.\\
For a video of $T$ frames containing $N$ people, each person $i$ at time step $t$ is represented as:

\begin{equation*}
    \mathbf{P}^i_t = \{\Phi^i_t, \Theta^i_t, \beta^i, \Gamma^i_t\}
\end{equation*}

where $\Phi^i_t\in\mathbb{R}^3$ is the global orientation, $\Theta^i_t\in\mathbb{R}^{22\times3}$ the body pose with 22 joint angles, $\beta^i\in\mathbb{R}^{11}$ the shape over all time steps $t$, where the $11^{th}$ value is the $\alpha$ interpolation weight, and $\Gamma^i_t\in\mathbb{R}^3$ the root translation.

The first step is to estimate each person's per-frame pose $\hat{\textbf{P}_t^i}$ and compute their unique identity track associations over all frames using a 3D tracking system, 4DHumans~\cite{goel2023humans}.

In a video, the net motion, \textit{i.e.}, a person's motion in the camera coordinates, depends both on the human's and camera's motion in the world frame. Therefore, the camera motion should be modeled in a correct way. Let $^c\mathbf{P}_t^i = \{^c\Phi_t^i, \Theta^i_t, \beta^i, ^c\Gamma_t^i \}$ the pose in the camera frame and $^w\mathbf{P}_t^i = \{^w\Phi_t^i, \Theta^i_t, \beta^i, ^w\Gamma_t^i \}$ the pose in the world with the same local pose $\hat{\Theta}_t^i$ and shape $\hat{\beta}^i$ parameters.

AionHMR-a uses DROID-SLAM~\cite{teed2021droid}, a SLAM system, to estimate the world-to-camera transform at each time $t$, $\{\hat{R}_t, \hat{T}_t\}$. This is essential to compute the relative camera motion between video frames. A human motion in the world prior is used to determine the camera scale $\alpha_c$ and people's global trajectories. The camera scale $\alpha_c$ is important to be estimated correctly to place the people in the world, so the human bodies and motion are plausible.

First, the global orientation and root translation in the world coordinate frame using the estimated camera transforms and camera-frame parameters are initialized. Camera scale is initialized in the value of $\alpha_c = 1$.

\begin{align*}
    ^w\Phi^i_t &= R_t^{-1c}\hat{\Phi}_t^i, && ^w{\Gamma}_t^i = R_t^{-1c}\hat{\Gamma}_t^i - \alpha_c R_t^{-1} T_t, \\
    \beta_i &= \hat{\beta}_i, && \Theta_t^i = \hat{\Theta}_t^i,
\end{align*}

The world frame joints are expressed as:

\begin{equation*}
    ^w\textbf{J}^i_t = \mathcal{M}(^w\Phi_t^i, \Theta^i_t, \beta^i) + ^w\Gamma^i_t
\end{equation*}

where $\mathcal{M}$ is the differentiable function that the SMPL~\cite{SMPL:2015} model uses to generate the mesh vertices and joints.

SLAHMR defines a 2D joint reprojection loss to align the projected 3D to 2D joints with the detected from ViTPose 2D keypoints $x^i_t$ that AionHMR-a also uses:

\begin{equation*}
    E_{\text{data}} = \sum_{i=1}^N\sum_{t=1}^T\psi_t^i\rho(\Pi_K(R_t\cdot^ w\bf{J}^i_t + \alpha T_t) - x_t^i)
\end{equation*}

where $\Pi_K(\begin{bmatrix}
    x_1 & x_2 & x_3
\end{bmatrix}^T) = K \begin{bmatrix}
    \frac{x_1}{x_3} & \frac{x_2}{x_3} & 1
\end{bmatrix}^T$ is perspective camera projection with camera intrinsics matrix $K \in \mathbb{R}^{2\times3}$, $\rho$ is the robust Geman-McClure function~\cite{barron2019generaladaptiverobustloss} and $\psi_t^i$ are the confidence scores of the detected 2D keypoints.

At this stage of the optimization, due to the under-constrained reprojection loss, the optimization is being held only to the global orientation and root translation $^w\Phi^i_t, ^w\Gamma^i_t$ of the human pose parameters:

\begin{equation*}
    \label{eq:global_orientation_root_translation}
    \min_{\{\{^w\Phi_t^i, ^w\Gamma^i_t\}^T_{t=1}\}^N_{i=1}}\lambda_{\text{data}}E_{\text{data}}
\end{equation*}

The optimization lasts 30 iterations with $\lambda_\text{data}=0.001$.

For the camera scale $\alpha_c$, human shape $\beta_i$ and body pose $\Theta_t^i$ optimization, additional priors about human movement in the world are used. This optimization stage smooths the transitions between poses in the world trajectories so that the displacements of the people are plausible. The prior of joint smoothness is defined as: 

\begin{equation*}
    E_{\text{smooth}} = \sum_i^N\sum_t^T\Vert\bf{J}_t^i - \bf{J}_{t+1}^i\Vert^2
\end{equation*}

The other priors concern the pose $E_{\text{pose}} = \sum_{i=2}^N\sum_{t=1}^T\Vert\zeta_t^i\Vert^2$ and the shape $E_\beta = \sum_i^N\Vert\beta^i\Vert^2$, where $\zeta_t^i\in\mathbb{R}^{32}$ represent the body pose parameters $\Theta_t^i$ in the latent space of the VPoser~\cite{SMPL-X:2019} model. The updated objective function to be minimized is the following:

\begin{equation*}
    \min_{\alpha,\{\{^w\bf{P}_t^i\}_{t=1}^T\}_{i=1}^N} \lambda_{\text{data}}E_{\text{data}}+\lambda_{\beta}E_{\beta} + \lambda_{\text{pose}}E_{\text{pose}} + \lambda_{\text{smooth}}E_{\text{smooth}}
\end{equation*}

The optimization is performed over 60 iterations using $\lambda_\text{smooth} = 5, \lambda_\beta=0.05$ and $\lambda_\text{pose}=0.04$.

\section{AionHMR-b Details}
\label{sec:aionb}

\subsection{Losses}

AionHMR-b uses different losses during the training, based on the ground-truth (or pseudo-ground-truth) annotations of the training datasets.

 If ground-truth SMPL-A~\cite{Patel:CVPR:2021} shape parameters $\beta^*$ and pose parameters $\theta^*$ are available, an MSE loss is used for the model predictions:

\begin{equation*}
    \mathcal{L}_{\texttt{smpl}} =\Vert \theta - \theta^*\Vert^2_2 + \Vert\beta-\beta^*\Vert^2_2
\end{equation*}

When the dataset provides accurate ground-truth 3D keypoints $X^*$, a L1 loss is added to penalize the distance from the predicted 3D keypoints $X$:

\begin{equation*}
    \mathcal{L}_{\texttt{kp3D}} = \Vert X - X^*\Vert_1
\end{equation*}

Similarly, if there are 2D keypoints annotations $x^*$, an L1 loss is used to penalize the projection of the predicted 3D keypoints $\pi(X)$:

\begin{equation*}
    \mathcal{L}_{\texttt{kp2D}} = \Vert\pi(X) - x^*\Vert_1
\end{equation*}

Finally, to get plausible 3D poses, a discriminator $D_k$ is trained for each factor of the body model, \ie, the body pose parameters $\theta_b$, the shape parameters $\beta$ and the per-part relative rotations $\theta_i$ with the generator loss expressed as:

\begin{equation*}
    \mathcal{L}_{\texttt{adv}} = \sum_k(D_k(\theta_b, \beta)-1)^2
\end{equation*}

\subsection{Architecture Details}

The AionHMR-b model that we train consists of one Vision Transformer~\cite{dosovitskiy2021imageworth16x16words} image encoder and a transformer decoder~\cite{NIPS2017_3f5ee243}. The ViT encoder is taken from the ViTPose model~\cite{xu2022vitpose}, which was pre-trained for the 2D joint detection task. It takes as input a $256 \times 192$ image and consists of 50 transformer layers. The encoder outputs $16 \times 12$ image tokens, each of dimension 1280. These tokens serve as the encoded representation of the input image for the decoder.

The transformer decoder has 6 layers, each with multi-head self-attention, multi-head cross-attention, and feed-forward blocks with layer normalization. It has a hidden dimension of 2048. Both the self-attention and cross-attention blocks use 8 heads, each with a dimension of 64. The feed-forward MLP has a hidden dimension of 1024.

For the SMPL-A parameters prediction, a 2048-dimensional learnable SMPL-A query token is fed into the transformer decoder. The decoder uses cross-attention to attend to the $16 \times 12$ image tokens from the ViT encoder. The output of the decoder is then passed through a linear layer to predict the final parameters. The output of the network consists of the pose ($\theta$), the shape ($\beta$), and the camera ($\pi$) parameters.

Table~\ref{tab:number_of_params} shows the number of trainable parameters for the backbone, the SMPL-A head and the discriminator, for a total of 671M parameters.

\begin{table}[h]
    \centering
    \scalebox{0.6}{
    \begin{tabular}{ccc}
    \toprule
    Name & Type & Number of Trainable Parameters \\
    \midrule
    backbone  & ViT & 630 M \\
    smpla\_head & SMPL-A Transformer Decoder Head & 39.5 M \\
    discriminator & Discriminator & 1.8 M\\
    \bottomrule
    \end{tabular}}
    \caption{Trainable Parameters for Model Components}
    \label{tab:number_of_params}
\end{table}

\subsection{Training Dataset}
For the initial training phase, we set the weights for the datasets as follows: the largest weight was assigned to our dataset (from SyRIP~\cite{huang2021infant} and Relative Human~\cite{BEV})(0.50), followed by AVA~\cite{gu2018ava}, AIC~\cite{wu2017ai}, and INSTA~\cite{kanazawa2019learning} datasets, each weighted at 0.10. The remaining datasets (Human3.6m~\cite{ionescu2013human3}, MPII~\cite{andriluka20142d}, COCO~\cite{lin2014microsoft}, and MPI-INF-3DHP~\cite{mehta2017monocular}) were each weighted at 0.05. We decided to assign the largest weight to our dataset in order for the model to learn the child and infant shape. For validation, we split the weight equally between ours (validation subset) and COCO-VAL (both at 0.50). For the subsequent fine-tuning phase, the weights were adjusted: our dataset weight was reduced to 0.3000, since we now focus on 3D pose training. The remaining eight training datasets were weighted equally at 0.0875. The validation dataset weights for fine-tuning remained unchanged; our validation subset and COCO-VAL both at 0.50.

\section{Action Preserving}
\label{sec:action}
\subsection{Description Generation}

To obtain detailed, consistent, and action-focused descriptions for both the original and reconstructed videos, we employed a Large Video Language Model (LVLM)~\cite{Qwen2.5-VL}. The following specific prompt was used to guide the LVLM to focus exclusively on human movement and ignore irrelevant details:

\noindent
\textbf{Prompt:} \texttt{Provide a detailed description of the actions of the human, initial position and the sequence of the movements. How the arms and the legs move, the orientation, as well as the head movements. Do not focus on the fingers and the facial expressions and do not describe the kid.}

\subsection{Description Similarity}

The core metric for action preservation is the semantic similarity between the two generated descriptions (original vs. reconstructed). We utilized a separate Large Language Model (LLM)~\cite{comanici2025gemini} specifically for semantic comparison. This model was prompted to output a percentage score based purely on the actions. The prompt used for the similarity comparison is provided below, emphasizing the exclusion of non-action-related details:

\noindent
\textbf{Prompt:} \texttt{Compare the two descriptions and give a percentage of similarity. Focus exclusively on the semantic similarity of the actions, movements, and implied behaviors of the human figure, while completely ignoring all details related to the setting (e.g., green mat, furniture, room description), clothing, and other appearance details.}

\subsubsection{Example Video Descriptions}

For illustration, we present two representative examples of the description pairs generated for the original video and its reconstructed counterpart, along with the similarity score assigned by the comparative LLM. The first exemplifies a case where, while the original video and the reconstructed video perform the same action from the same child, the description similarity is low. On the other hand, the second example presents two descriptions where the LLM assigned a semantic similarity score of 100\%. \\

\noindent
\textbf{Similarity: 40\%}

\begin{itemize}
   \item \textbf{Original}: \texttt{The individual is standing on a green mat with colorful house patterns in a room that appears to be set up for play or learning activities. The person is wearing a dark blue t-shirt with white text and light gray shorts. Their arms are initially at their sides, and they are facing forward.In the first frame, the person's right arm moves slightly upward, with the hand coming close to the shoulder level. The left arm remains relatively still, hanging by the side. The head is oriented straight ahead, looking directly at something off-camera to the right. In the second frame, the person's right arm continues to move upward, now reaching closer to the chest area. The left arm remains still. The head maintains its forward orientation, still focused on the same point off-camera. The sequence of movements suggests a gesture or action being performed, possibly in response to an instruction or interaction with someone or something out of view. The individual's body remains mostly stationary.}
    \item \textbf{AionHMR}: \texttt{The human figure is standing on a green mat with a colorful pattern that resembles a cityscape. The figure's arms are initially at its sides, but it begins to move them in a rhythmic manner, clapping them together in front of its chest. The arms move up and down in a synchronized motion, creating a clapping rhythm. The legs remain stationary throughout this action, maintaining a straight posture. The head remains oriented forward, facing the direction of the camera, and there are no noticeable changes in the orientation or movement of the head. The overall posture of the figure suggests a calm and controlled movement, focusing primarily on the clapping action.}
 
\end{itemize}

\begin{figure*}[ht!]
    \centering
    \includegraphics[width=0.95\linewidth]{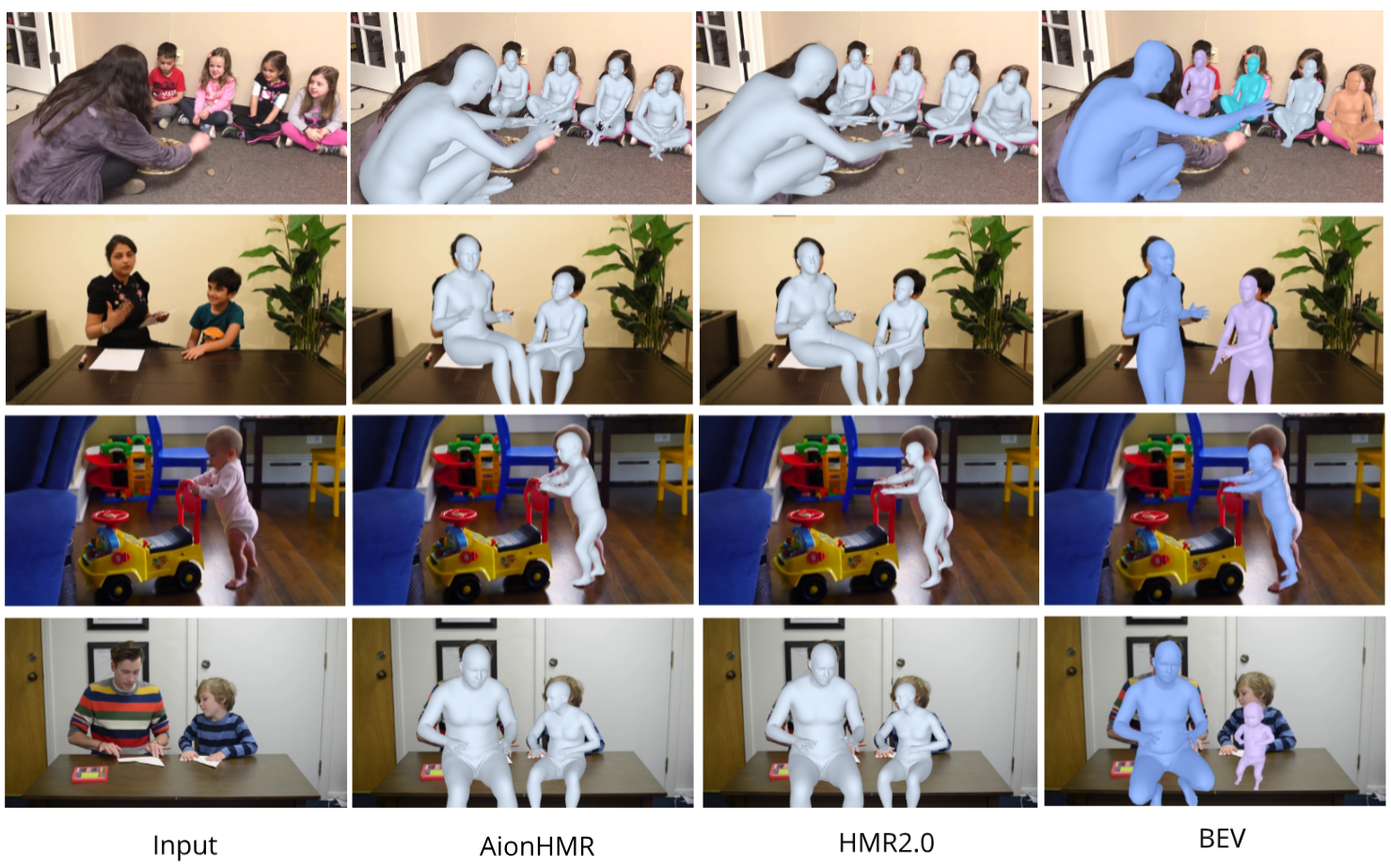}
    \caption{\textbf{AionHMR and Baselines Comparisons Examples.} From left to right: Original Image, AionHMR, HMR2.0 and BEV. Compared to HMR2.0 and BEV, AionHMR provides the most accurate estimation of the child's shape, while effectively estimating the pose.}
    \label{fig:comps_suppl}
\end{figure*}

\noindent
\textbf{Similarity: 100\%}
\begin{itemize}
 \item \textbf{Original} \texttt{The human in the video is a young girl standing still with her arms relaxed at her sides. Her body is facing forward, and her head is oriented straight ahead. As the sequence progresses, the girl begins to walk across the room. She takes small, steady steps, moving her feet one after the other. Her arms swing naturally by her sides as she walks. Her head remains level and oriented forward, following the direction of her gaze as she moves. The girl continues walking across the room, maintaining a consistent pace and posture throughout the movement.}
    \item \textbf{AionHMR}: \texttt{The human figure's arms are initially by its sides, and it appears to be facing forward. The legs are straight, and the feet are flat on the ground. As the sequence progresses: 1. The figure begins to walk forward, taking small steps. 2. The arms swing naturally at the sides as the figure moves. 3. The head remains oriented forward, following the direction of movement. 4. The legs continue to move in a walking motion, with each step bringing one foot forward and the other foot following behind. 5. The overall posture remains upright and balanced throughout the movement. The figure maintains a consistent pace and direction, moving steadily across the mat.}
   
\end{itemize}

\section{Qualitative Results}
\label{sec:qual}

\begin{figure*}
    \centering
    \includegraphics[width=0.78\linewidth]{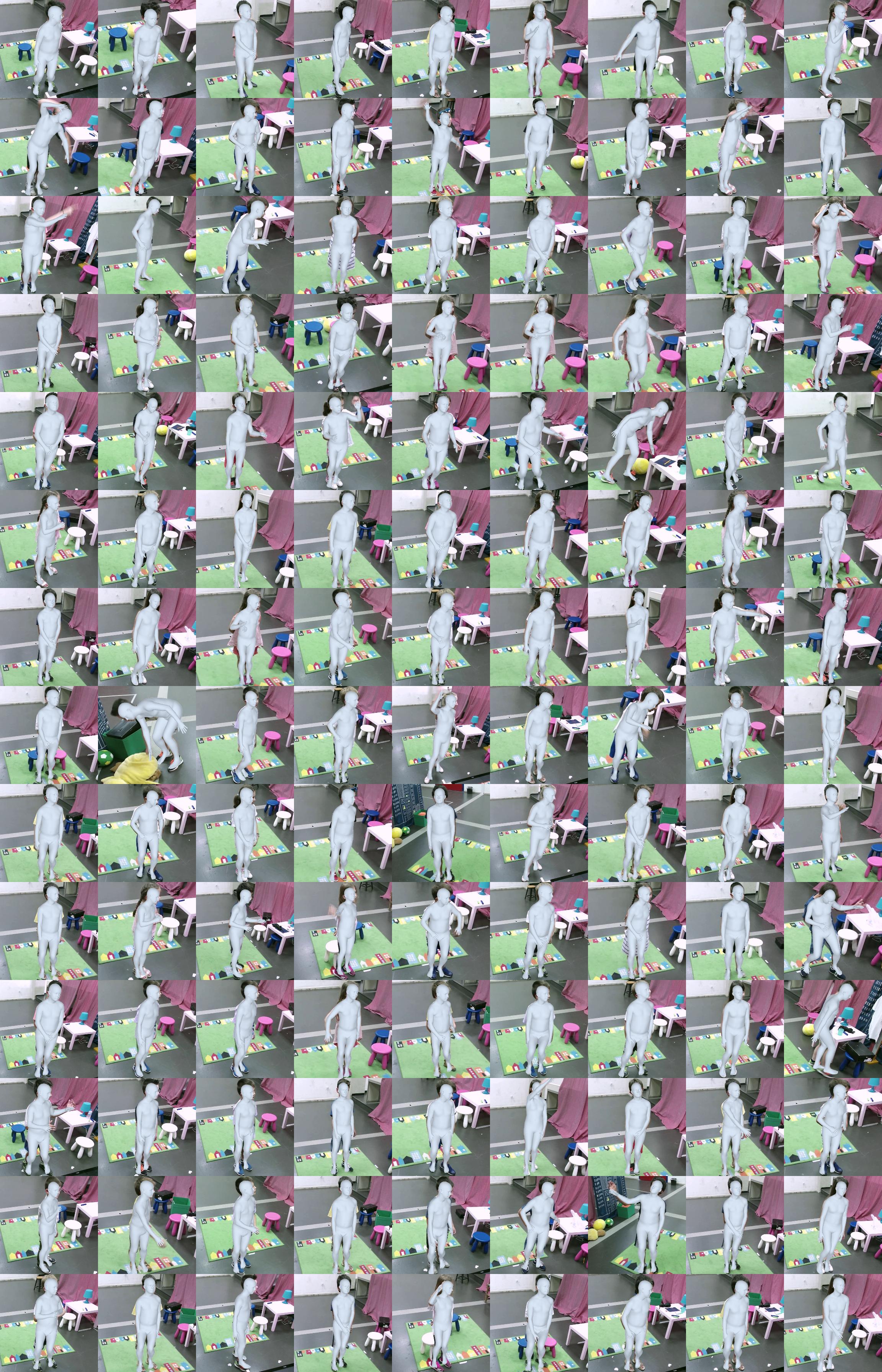}
    \caption{\textbf{Qualitative results from the 3D-BabyRobot Dataset}. The figure showcases frames from our newly released 3D-BabyRobot Dataset (over 4M frames of child-robot interaction). The dataset provides 3D reconstructions of children from AionHMR (instead of the original children or raw identity data). These examples serve as qualitative results, demonstrating AionHMR's strong capability for accurate 3D shape and pose estimation.}
    \label{fig:3d-baby-one-page}
\end{figure*}

In Figure~\ref{fig:comps_suppl} we present some comparisons of AionHMR and two baselines, HMR2.0 and BEV. 
Figure~\ref{fig:3d-baby-one-page} provides an extensive sample from the 3D-BabyRobot Dataset, which also serves as a qualitative assessment of AionHMR.

\end{document}